\documentclass[sigconf]{acmart}
\AtBeginDocument{%
  }

\usepackage[table]{xcolor}
\usepackage{pifont}
\usepackage{threeparttable}
\usepackage{graphicx}
\usepackage{tabularx}
\newcolumntype{Y}{>{\raggedright\arraybackslash}X}
\usepackage[utf8]{inputenc}
\usepackage{geometry}
\usepackage{booktabs}   
\usepackage{graphicx}   
\usepackage{hyperref}
\usepackage{xcolor}
\usepackage{array}
\usepackage{makecell} 
\newcolumntype{C}[1]{>{\hsize=#1\hsize\centering\arraybackslash}X}

\usepackage{siunitx} 

\newcommand{\greencheck}{\textcolor{green!60!black}{\ding{51}}}
\newcommand{\redx}{\textcolor{red!40!gray}{\ding{55}}}

\hypersetup{
    colorlinks=true,
    linkcolor=blue,
    urlcolor=cyan,
    breaklinks=true
}

\usepackage{tcolorbox}
\usepackage{hyperref}
\tcbuselibrary{skins, breakable}

\setcopyright{none}
\renewcommand\footnotetextcopyrightpermission[1]{} 
\begin{document}

\title{LiveMedBench: A Contamination-Free Medical Benchmark for LLMs with Automated Rubric Evaluation}

\author{Zhiling Yan}
\affiliation{%
  \institution{Lehigh University}
  \city{Bethlehem, PA}
  \country{USA}
}
\author{Dingjie Song}
\affiliation{%
  \institution{Lehigh University}
  \city{Bethlehem, PA}
  \country{USA}
}
\author{Zhe Fang}
\affiliation{%
  \institution{Harvard University}
  \city{Boston, MA}
  \country{USA}
}
\author{Yisheng Ji}
\affiliation{%
  \institution{Imperial College London}
  \city{London}
  \country{United Kingdom}
}
\author{Xiang Li}
\affiliation{%
  \institution{Massachusetts General Hospital and Harvard Medical School}
  \city{Boston, MA}
  \country{USA}
}
\author{Quanzheng Li}
\affiliation{%
  \institution{Massachusetts General Hospital and Harvard Medical School}
  \city{Boston, MA}
  \country{USA}
}
\author{Lichao Sun}
\authornote{Corresponding author.}
\affiliation{%
  \institution{Lehigh University}
  \city{Bethlehem, PA}
  \country{USA}
}
\renewcommand{\shortauthors}{Zhiling Yan et al.}

\begin{abstract}
The deployment of Large Language Models (LLMs) in high-stakes clinical settings demands rigorous and reliable evaluation. However, existing medical benchmarks remain static, suffering from two critical limitations: (1) data contamination, where test sets inadvertently leak into training corpora, leading to inflated performance estimates; and (2) temporal misalignment, failing to capture the rapid evolution of medical knowledge. Furthermore, current evaluation metrics for open-ended clinical reasoning often rely on either shallow lexical overlap (e.g., ROUGE) or subjective LLM-as-a-Judge scoring, both inadequate for verifying clinical correctness.
To bridge these gaps, we introduce \textbf{LiveMedBench}, a continuously updated, contamination-free, and rubric-based benchmark that weekly harvests real-world clinical cases from online medical communities, ensuring strict temporal separation from model training data. We propose a Multi-Agent Clinical Curation Framework that filters raw data noise and validates clinical integrity against evidence-based medical principles. For evaluation, we develop an Automated Rubric-based Evaluation Framework that decomposes physician responses into granular, case-specific criteria, achieving substantially stronger alignment with expert physicians than LLM-as-a-Judge.
To date, LiveMedBench comprises 2,756 real-world cases spanning 38 medical specialties and multiple languages, paired with 16,702 unique evaluation criteria. Extensive evaluation of 38 LLMs reveals that even the best-performing model achieves only 39.2\%, and 84\% of models exhibit performance degradation on post-cutoff cases, confirming pervasive data contamination risks. Error analysis further identifies contextual application---not factual knowledge---as the dominant bottleneck, with 35--48\% of failures stemming from the inability to tailor medical knowledge to patient-specific constraints. The code and data are available at \href{https://github.com/ZhilingYan/LiveMedBench/}{LiveMedBench}.

\end{abstract}


\vspace{-10pt}


\maketitle

\section{Introduction}


Large Language Models (LLMs) are increasingly applied in healthcare, assisting with tasks ranging from answering patient inquiries to supporting clinical decision-making~\cite{singhal2023clinical}. Deploying AI in high-stakes clinical settings requires that these models demonstrate robust performance across diverse, real-world scenarios~\cite{nori2023capabilities}.


\begin{figure}[t]
  \centering
  \includegraphics[width=\columnwidth]{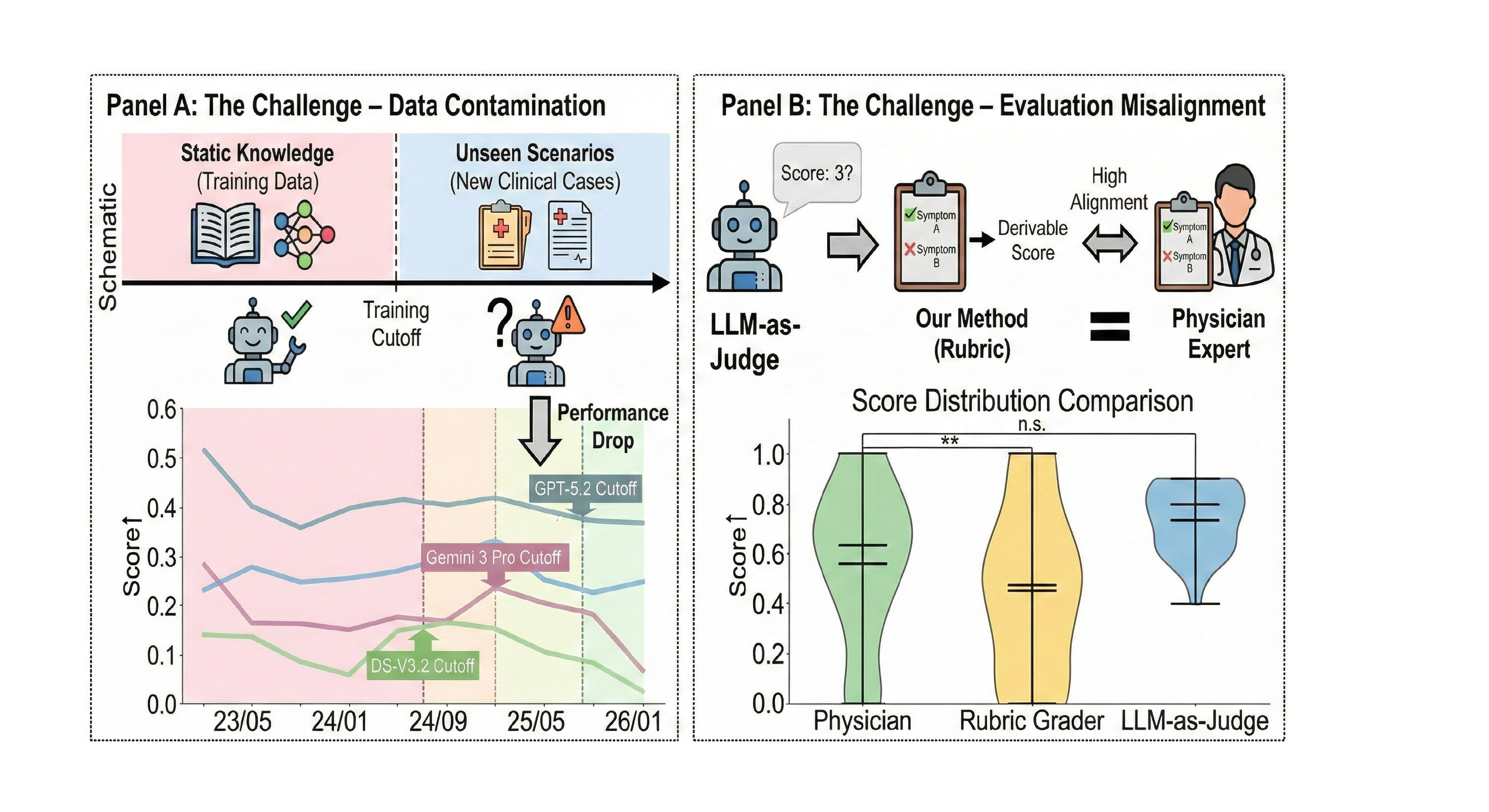}
  \vspace{-2mm}
  \vspace{-2mm}
  \caption{(A) Temporal degradation. Model performance consistently declines on clinical cases that post-date their training knowledge cutoffs, highlighting the risk of data contamination. (B) Evaluation alignment. Our proposed Automated Rubric-based Evaluation Framework aligns better with physician experts compared to LLM-as-a-Judge.}
  \label{fig:intro-example}
  \vspace{-10pt}
\end{figure}

While benchmarks serve as essential standards for tracking progress, existing medical benchmarks face substantial obstacles in both data construction and evaluation.

First, on the data side, most benchmarks are static~\cite{singhal2023clinical}, leading to two issues. (i) They become temporally misaligned with rapidly evolving medical knowledge. Clinical guidelines and health policies are frequently updated, and new diseases and public health emergencies can emerge unexpectedly~\cite{leong2025generalist}. As a result, static test sets risk becoming increasingly misaligned with the evolving standards of real-world clinical practice~\cite{cheshmehzangi2025crisis}. (ii) They are vulnerable to data contamination. Because LLMs are trained on massive and often opaque corpora, publicly released benchmarks may inadvertently appear in training data. Consequently, strong scores may reflect memorization rather than genuine clinical reasoning, undermining credibility of  evaluation~\cite{magar2022data}.

Second, on the evaluation side, commonly used metrics remain inadequate for open-ended clinical reasoning. Lexical-overlap metrics (e.g., BLEU) fail to capture semantic and clinical correctness~\cite{lin2004rouge,papineni2002bleu}, and multiple-choice formats are poorly aligned with the open-ended and dynamic nature of real-world clinical workflows~\cite{jin2019medqa}. Recent work has adopted LLM-as-a-Judge~\cite{zheng2024judging} for holistic scoring, while they rely on implicit model intuition and typically lack explicit, fine-grained criteria that support objective verification~\cite{healthbench}.


While recent works attempt to address these challenges, they fall short. Approaches utilizing LLM prompts to modify data, e.g. dynamic benchmarks~\cite{zhang2025inflated,song2024both}, cannot update  underlying seed cases. They fail to reflect real-world medical evolution and risk introducing model-generated biases. Conversely, benchmarks like MedArena~\cite{medarena} and HealthBench~\cite{healthbench} depend on extensive human annotation, making them prohibitively expensive and difficult to scale.




To address these challenges, we introduce \textbf{LiveMedBench}, a multilingual, multi-specialty, rubric-based benchmark that comprises 2,756 real-world cases paired with 16,702 unique evaluation criteria.
To build LiveMedBench, we develop a data pipeline---the \textbf{Multi-Agent Clinical Curation Framework}---that continuously harvests new clinical cases (weekly) from active online medical communities, focusing on scenarios resolved by verified physicians. The pipeline coordinates multiple agents with Retrieval-Augmented Generation (RAG)~\cite{lewis2020retrieval, medrag} to cross-check case details against authoritative medical evidence, thereby validating clinical plausibility and enforcing consistency with established standards.
For evaluation, we propose an \textbf{Automated Rubric-based Evaluation Framework} that converts physician responses into case-specific rubrics consisting of self-contained, objective criteria and uses them to grade open-ended model outputs, achieving strong alignment with human experts (Figure~\ref{fig:intro-example} - Panel B).

%

We evaluate 38 LLMs spanning both proprietary and open-source models across general-purpose and medically specialized     
architectures. As illustrated in Figure~\ref{fig:intro-example} - Panel A, performance diverges markedly around the knowledge cutoff date. This degradation is consistent with temporal misalignment and potential exposure to earlier cases.
Notably, even the state-of-the-art GPT-5.2 achieves only 39.2\%, while 84\% of evaluated models exhibit significant performance drops on post-cutoff cases, underscoring the pervasive risk of data contamination. Our error analysis identifies a critical failure mode: the dominant bottleneck is contextual application—the inability to synthesize medical facts with patient-specific constraints (35–48\% of errors). Furthermore, retrieval-based knowledge injection recovers much of this performance loss, confirming that contemporary failures stem from knowledge gaps rather than innate reasoning deficits.
In summary, our main contributions are:
\begin{enumerate}
  \item We introduce LiveMedBench, a novel benchmark updated weekly to mitigate data contamination and knowledge obsolescence. To date, the dataset comprises 2,756 real-world cases spanning 38 specialties and multiple languages.
  \item We propose a Multi-Agent Clinical Curation Framework that leverages automated evidence verification to align benchmark data with authoritative medical standards, ensuring high clinical integrity.
  \item We develop an Automated Rubric-based Evaluation Framework that enables scalable, fine-grained assessment by decomposing expert responses into 16,702 unique, case-specific criteria.
  \item We provide a rigorous evaluation of 38 LLMs, systematically contrasting proprietary and open-source models across general and medical domains to characterize the current landscape of clinical AI.
\end{enumerate}


\section{Related Work}

\subsection{Medical Benchmarks}
Medical LLM evaluation has evolved from static knowledge tests like MedQA \cite{jin2019medqa} and PubMedQA \cite{jin2019pubmedqa} toward open-ended reasoning. Subsequent work introduced free-response generation \cite{singhal2023clinical}, multimodal inputs \cite{li2023llavamed}, and multilingual settings \cite{wang2024cmb}. This shift necessitated a transition from lexical-overlap metrics (e.g., ROUGE) to LLM-as-a-judge \cite{zheng2024judging} and structured rubric-based assessment. HealthBench \cite{healthbench} pioneered the use of case-specific rubrics to evaluate clinical safety and communication. Recent efforts further emphasize complex diagnostic reasoning and multi-step planning \cite{tu2025diagnostic, mcduff2025differential}. LiveMedBench builds on these trends by integrating automated, scalable rubric generation with continuously refreshed real-world cases.

\vspace{-5pt}
\subsection{Dynamic and Live Benchmarks}
Static benchmarks are increasingly vulnerable to data contamination and temporal misalignment \cite{magar2022data}. To ensure contamination free evaluation, frameworks like LiveBench \cite{livebench} and LiveCodeBench \cite{livecodebench} harvest fresh content from recent news and coding competitions. In the medical domain, DyRe-Me \cite{zhang2025inflated} and MedPerturb \cite{medperturb} mitigate contamination by perturbing existing seed cases. However, these methods often rely on objective metrics (e.g., unit tests) unsuitable for clinical ambiguity. LiveMedBench adapts the live paradigm to high-stakes medicine, employing a rigorous multi-agent system to validate clinical integrity and assess dimensions like safety and empathy that logic-only benchmarks overlook.

\begin{figure*}[t]
  \centering
  \includegraphics[width=\textwidth]{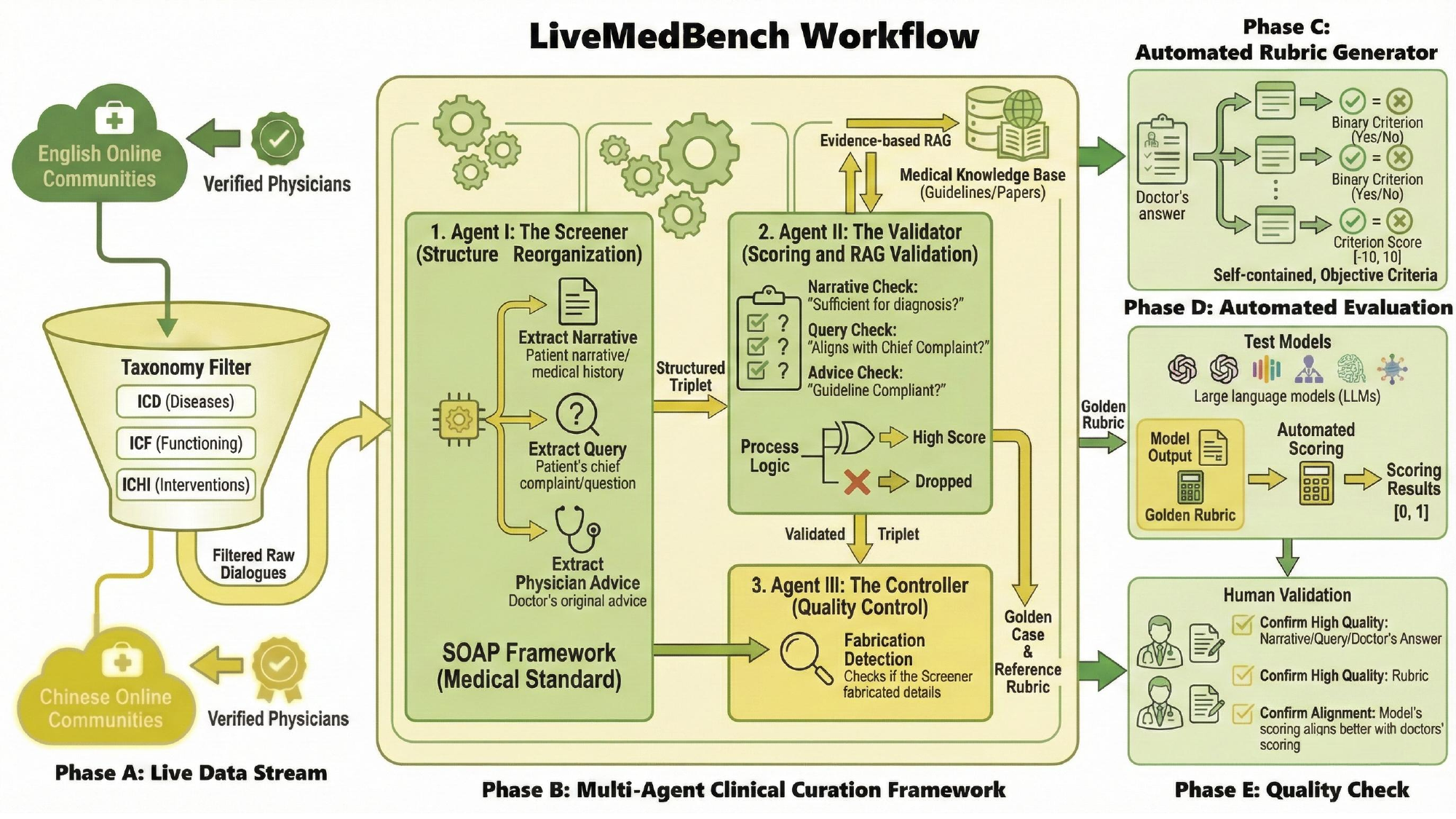}
  \vspace{-2mm}
  \vspace{-2mm}
  \caption{Overview of the LiveMedBench framework. The pipeline consists of five phases: (a) Continuous mining of bilingual clinical data from verified online communities; (b) A Multi-Agent Curation Framework (Screener, Validator, Controller) that structures and validates data against medical guidelines; (c) Automated generation of case-specific evaluation rubrics; (d) Objective evaluation of LLMs using the generated rubrics; and (e) Rigorous human quality assurance to ensure clinical alignment.}
  \label{fig:workflow}
  \vspace{-2mm}
\end{figure*}

\vspace{-5pt}
\section{LiveMedBench}
\label{sec:livemedbench}

In this section, we describe the end-to-end construction and evaluation pipeline of LiveMedBench (Figure~\ref{fig:workflow}). We first introduce data collection and filtering from online medical communities (\S\ref{sec:data-collection}), then present our Multi-Agent Clinical Curation Framework for transforming noisy threads into structured, clinically grounded cases (\S\ref{sec:curation-framework}), describe our Automated Rubric-based Evaluation
Framework, which distills expert physician responses into case-specific criteria for scalable, objective evaluation (\S\ref{sec:rubric-generator}), and finally summarize the resulting dataset statistics (\S\ref{sec:data-statistics}).

\vspace{-5pt}
\subsection{Data Collection}
\label{sec:data-collection}

\paragraph{Platform Selection.}
To ensure the reliability and clinical relevance of our source data, we selected platforms that are (i) \textbf{high activity}, with new user-generated posts and discussions on a weekly basis; (ii) \textbf{institutionally endorsed}, i.e., affiliated with reputable medical organizations (e.g., the National Comprehensive Cancer Network (NCCN) and the Chinese Medical Doctor Association) or major tertiary hospitals; and (iii) \textbf{professionally verified}, i.e., hosting a large pool of verified physicians (ranging from 4,500+ specialists to 2M+ registered professionals). Based on these criteria, we harvest raw clinical discussions from four premier online medical communities: iCliniq\footnote{\url{https://www.icliniq.com}}, SDN\footnote{\url{https://www.studentdoctor.net}}, DXY\footnote{\url{https://dxy.com}}, and Medlive\footnote{\url{https://www.medlive.cn}}. Details of these communities are provided in Appendix~\ref{app:dataset}.

\paragraph{Post Filtering.}
To improve data quality, we implement a filtering protocol following~\cite{singhal2023clinical}. We retain only posts and discussions that were published on or after January 1, 2023 (to reduce contamination risk and align with current medical knowledge), have clinically relevant tags (excluding non-clinical topics such as career planning or licensing exams), are text-only (threads requiring images, videos, or audio are discarded), contain fewer than 50\% non-alphanumeric characters, are written in English or Chinese, include at least one keyword from the ICD~\cite{who_icd}, ICF~\cite{who_icf}, or ICHI~\cite{who_ichi} codebooks, are not duplicates of previously indexed entries, and include at least one verified physician response.



\vspace{-5pt}
\subsection{Multi-Agent Clinical Curation}
\label{sec:curation-framework}

To transform unstructured online discussions into high-fidelity clinical cases, we propose a Multi-Agent Clinical Curation Framework. This framework simulates a collaborative expert validation mechanism, employing three specialized agents that operate in a hierarchical workflow to ensure structural integrity, clinical plausibility, and evidence alignment. Let the raw input be a discussion thread $T = \{m_1, m_2, ..., m_k\}$, where $m_i$ represents the $i$-th message in the thread. The objective is to convert raw discussions into triplets: $D = \{Narrative (N), Query (Q), Advice (A)\}$.

\subsubsection{The Screener}
The Screener standardizes each raw discussion thread $T$ into a structured clinical representation following the SOAP (Subjective, Objective, Assessment, Plan) framework~\cite{soap_guide}. It outputs a triplet
$$ D_{screener} = \{L_N, Q, L_A\}, $$
where $L_N$ is the set of atomic patient narrative entities (Subjective/Objective), $Q$ is the primary user query, and $L_A$ is the set of actionable physician advice elements (Assessment/Plan). Concretely, the Screener performs:
\noindent\textbf{(i) Narrative aggregation ($L_N$):} extract and merge patient-provided clinical facts from $T$.
\noindent\textbf{(ii) Query extraction ($Q$):} identify the explicit question/request for medical guidance.
\noindent\textbf{(iii) Advice extraction ($L_A$):} extract actionable diagnostic or management recommendations from verified physician replies.
A candidate is forwarded only if all components are present: $Forward(Case) \iff (L_N \neq \emptyset) \land (Q \neq \emptyset) \land (L_A \neq \emptyset)$

\subsubsection{The Validator}

The Validator assesses the quality and factual consistency of the structured triplet $D_{screener} = \{L_N, Q, L_A\}$ and outputs a validation vector 
$$V = \{S_{cc}, S_{inf}, S_{align}\}.$$
\noindent\textbf{(i) Query validity ($S_{cc}$):} a binary score indicating whether $Q$ is a clinically meaningful chief complaint (CC) as defined by Ely's taxonomy~\cite{ely_taxonomy}.
\noindent\textbf{(ii) Narrative sufficiency ($S_{inf}$):} the fraction of required clinical information items (derived from $Q$) that are supported (entailed) by the patient narrative $L_N$, inspired by~\cite{es2024ragas}. Concretely, we first derive a minimal checklist $S_{req}$ of atomic requirements needed to answer $Q$ (e.g., onset, severity, red flags), and then compute $S_{inf}$ as the proportion of items in $S_{req}$ that can be verified from $L_N$.
\noindent\textbf{(iii) Evidence alignment ($S_{align}$):} an evidence-based validity score for physician advice $L_A$ computed by checking each advice element against retrieved medical evidence (Clinical Guidelines, PubMed, StatPearls, etc.)~\cite{meditron,medrag}. Following MedRAG~\cite{medrag}, we retrieve top-$k$ evidence snippets using a MedCPT retriever~\cite{medcpt} and score each advice element with $v(a_j) \in \{1.0, 0.5, 0.0\}$ (supported/neutral/contradicted). We apply a veto strategy: if any $a_j$ is contradicted, the entire case is discarded.
A candidate case $D$ is retained \textbf{if and only if} $Keep(D) \iff (S_{cc} = 1) \land (S_{inf} > \tau_{inf}) \land (S_{align} > \tau_{align})$, where $\tau_{inf}=0.5$ and $\tau_{align}=0.5$ are predefined thresholds.

\subsubsection{The Controller}

The Controller accepts the valid candidate $D_{candidate} = \{L_N, Q, L_A\}$ from the Validator and performs a final audit before generating the canonical natural language output. It verifies that every atomic element in $L_N \cup L_A$ is explicitly supported by the original thread $T$:
$$Keep(D) \iff \forall x \in (L_N \cup L_A), \text{Supported}(x, T).$$
Any unsupported or fabricated element leads to discarding the case to prevent synthetic contamination. Cases that pass are synthesized into coherent narrative ($N$) and advice ($A$), yielding the final triplet $D = \{N, Q, A\}$.

\subsection{Automated Rubric-based Evaluation}
\label{sec:rubric-generator}

To facilitate scalable and objective assessment, we propose an \textbf{Automated Rubric-based Evaluation Framework}. This pipeline consists of two distinct modules: a \textit{Rubric Generator} that converts physician responses into granular grading criteria, and a \textit{Rubric-based Grader} that assesses model outputs against these criteria.

\subsubsection{Automated Rubric Generator}
Following the taxonomy of \citet{healthbench}, the generator transforms the verified advice list $L_A$ into a structured rubric $R$. We adapt five of \citet{healthbench}'s seven themes to assign each case a dominant \textbf{Theme}---\textit{Emergency Referrals}, \textit{Context-Seeking}, \textit{Expertise-Tailored Communication}, \textit{Responding under Uncertainty}, or \textit{Response Depth}. This theme serves as a contextual filter for the subsequent generation steps:

\noindent \textbf{Step 1: Theme-Guided Fact Extraction.} The agent filters $L_A$ to retain only medical facts and instructions relevant to the assigned Theme. For instance, if the Theme is ``Emergency Referrals,'' long-term dietary advice is discarded to focus strictly on triage accuracy.
    
\noindent \textbf{Step 2: Bipolar Criterion Formulation.} Extracted facts are converted into binary criteria with dual polarity. \textbf{Positive criteria} reward the correct inclusion of key facts (e.g., ``Does the model identify the likely cause as Norovirus?''), while \textbf{negative criteria} penalize hallucinations or contradictions (e.g., ``Does the model incorrectly suggest antibiotics for a viral infection?'').
    
\noindent \textbf{Step 3: Axis Assignment and Weighting.} Each criterion is labeled with an evaluation \textbf{Axis} (\textit{Accuracy}, \textit{Completeness}, \textit{Communication Quality}, \textit{Context Awareness}, or \textit{Safety}) and assigned a weight $w_j \in [-10, 10]$. Positive weights denote reward criteria, while negative weights flag penalties. Extreme weights (e.g., $w_j = \pm 10$) represent life-critical details where errors could compromise patient safety.

\subsubsection{Rubric-based Grader}
Following the generation of the case-specific rubric $R = \{(c_1, w_1), \dots, (c_m, w_m)\}$, the Rubric-based Grader systematically assesses the model's response by determining the satisfaction of each atomic criterion.
We employ a model grader to independently verify whether each criterion $c_j$ is \textit{met} or \textit{not met} based on the model's output.

The final per-case score is calculated by summing the weights of satisfied criteria, normalized by the maximum possible positive score:
\begin{equation}
    Score = \text{clip}\!\left( \frac{\sum_{j=1}^{m} w_j \cdot \mathbb{I}(\text{Model} \models c_j)}{\sum_{k \in \{p \mid w_p > 0\}} w_k},\; 0,\; 1 \right)
\end{equation}
where $\mathbb{I}(\cdot)$ is the indicator function equal to 1 if the criterion is met, and 0 otherwise. The denominator represents the total score achievable if all positive criteria are met. The clipping function ensures the score remains within $[0, 1]$ even if substantial penalties (negative weights) occur. A model's overall LiveMedBench score is derived from the mean of these case-level scores.

\begin{figure*}[t]
  \centering
  \includegraphics[width=\textwidth]{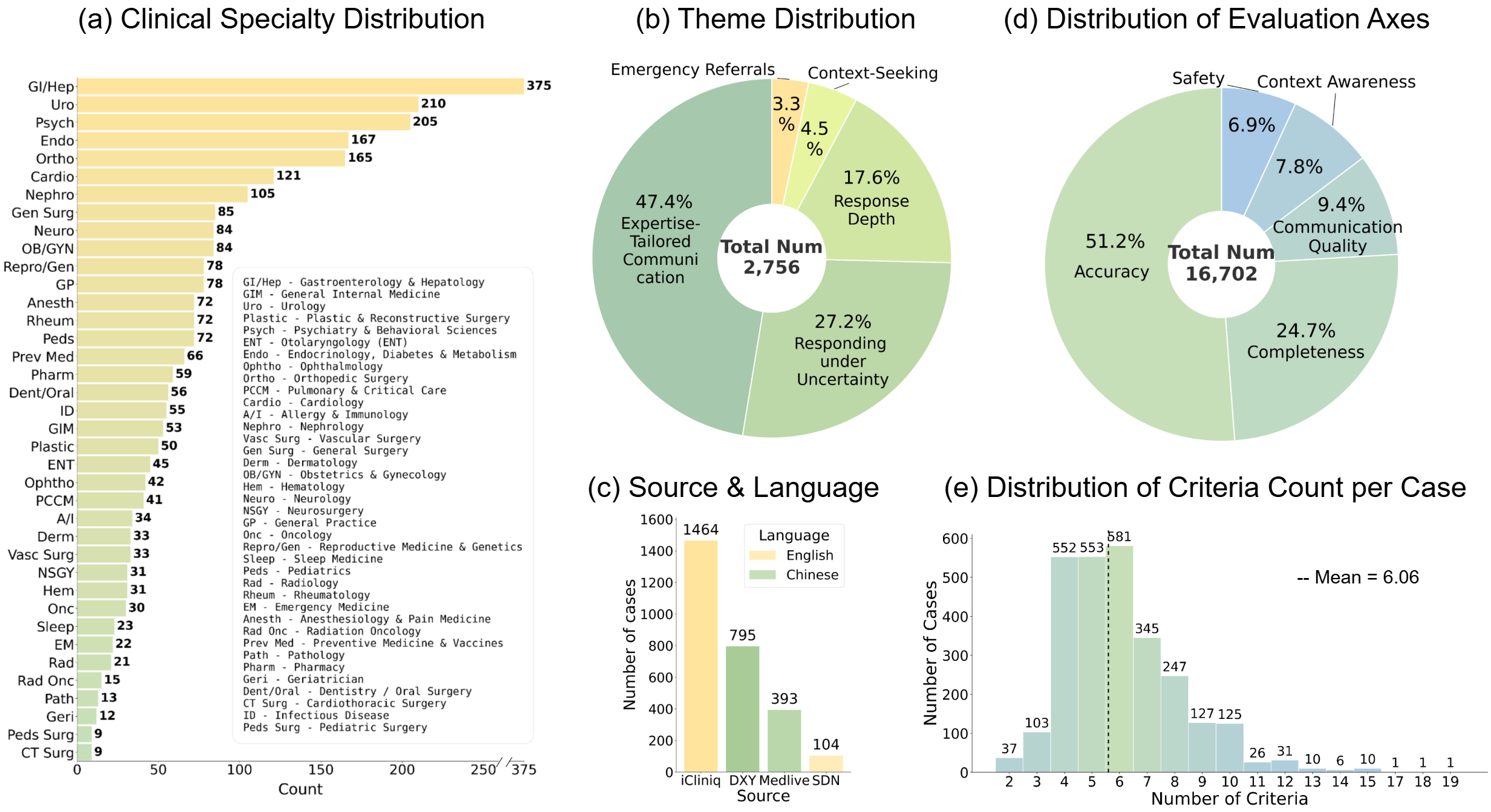}
  \vspace{-2mm}
  \vspace{-2mm}
  \vspace{-2mm}
  \vspace{-2mm}
  \caption{Data statistics of LiveMedBench. The figure illustrates the comprehensive distribution of (a) 38 clinical specialties, (b) five behavioral themes, (c) data sources and languages , (d) evaluation axes, and (e) the number of grading criteria per case (Mean=6.06).}
  \vspace{-2mm}
  \vspace{-2mm}
  \label{fig:statistics}
\end{figure*}


\subsection{Data Statistics}                                                                                 
\label{sec:data-statistics}                                                                                  
                                                                                                           
LiveMedBench currently comprises 2,756 clinical cases and 16,702 unique rubric criteria. As illustrated in Figure~\ref{fig:statistics}, the dataset spans 38 clinical specialties with a naturally long-tailed distribution reflecting real-world prevalence, 1,568 English and 1,188 Chinese cases sourced from four platforms, and five behavioral themes led by Expertise-Tailored Communication (47.4\%). The rubric criteria cover five evaluation axes---predominantly Accuracy (51.2\%) and Completeness (24.7\%)---with an average of 6.06 criteria per case (range: 2--19). A detailed breakdown is provided in Appendix~\ref{app:data-statistics}.

\section{Experiments}

\subsection{Experimental Setup}




\textbf{Models Evaluated}. We conducted a comprehensive evaluation across 38 LLMs, categorized into three distinct groups to ensure a holistic assessment: (1) Proprietary general-purpose models, including the GPT-series~\cite{gpt_series} and Gemini-series~\cite{gemini_series}; (2) Open-source general-purpose models, such as the DeepSeek-series~\cite{deepseek_series} and Qwen-series~\cite{qwen_series,team2024qwen2}; and (3) Medical-specific models, including the Baichuan-series~\cite{baichuan_series,baichuan-m3} and MedGemma-series~\cite{medgemma}. All models were evaluated in a zero-shot setting to assess their innate clinical capabilities without task-specific learning. For reproducibility, we set the temperature to 0. Detailed model specifications are provided in Appendix~\ref{app:model-specs}.

\textbf{Implementation Details}. We detail the specific models employed in our frameworks. In the \textit{Multi-agent Clinical Curation} phase, both the \textit{Screener} and \textit{Controller} agents utilized \texttt{Qwen/Qwen3-4B-Instruct-2507} for efficient processing, while the \textit{Validator} agent employed \texttt{openai/gpt-oss-120b} to ensure rigorous verification. For the \textit{Automated Rubric-based Evaluation}, the \textit{Rubric Generator} was powered by \texttt{Qwen/Qwen3-4B-Instruct-2507}, while \textit{Rubric-based Grader} uses \texttt{gpt-4.1-2025-04-14} following~\citet{healthbench}.

\textbf{Evaluation Metric}. For each case, the grader evaluates the model's response against the specific rubric criteria generated by the Rubric Generator using a binary judgment (Met vs. Not Met). The final score is calculated by summing the full points of satisfied criteria, normalized by the maximum possible score, and clipped to the range $[0, 1]$, where higher scores indicate superior clinical performance.

\textbf{Human Validation}. To validate the data quality and the reliability of our evaluation framework, we conducted a rigorous human evaluation with two physicians proficient in both English and Chinese. We randomly sampled 50 cases (comprising 292 unique criteria) and performed assessment across three dimensions:
\noindent \textbf{1. Data Quality:} Physicians rated the clinical coherence of the patient narrative/query and the quality of the reference physician advice on a 3-point scale ($0-2$), where scores $\ge 1$ denote high quality and agreement.
\noindent \textbf{2. Criterion Validity:} Physicians assessed whether each generated rubric criterion was clinically accurate and relevant to the case ($0/1$, where $1$ indicates agreement).
\noindent \textbf{3. Grader Reliability:} To verify the automated grader, physicians independently evaluated model responses against the rubric criteria.

We calculated the inter-rater reliability between physicians using Gwet's AC1~\cite{gwet2008}. To quantify the alignment between the model grader and human experts, we employed two statistical measures: (1) the Macro F1 score for criterion-level agreement, following \citet{healthbench}; and (2) the Pearson correlation coefficient for case-level scoring alignment. Further details on the human evaluation protocol are available in Appendix~\ref{app:human-study}.

\subsection{Overall Model Performance}

\begin{figure*}[t]
  \centering
  \includegraphics[width=\textwidth]{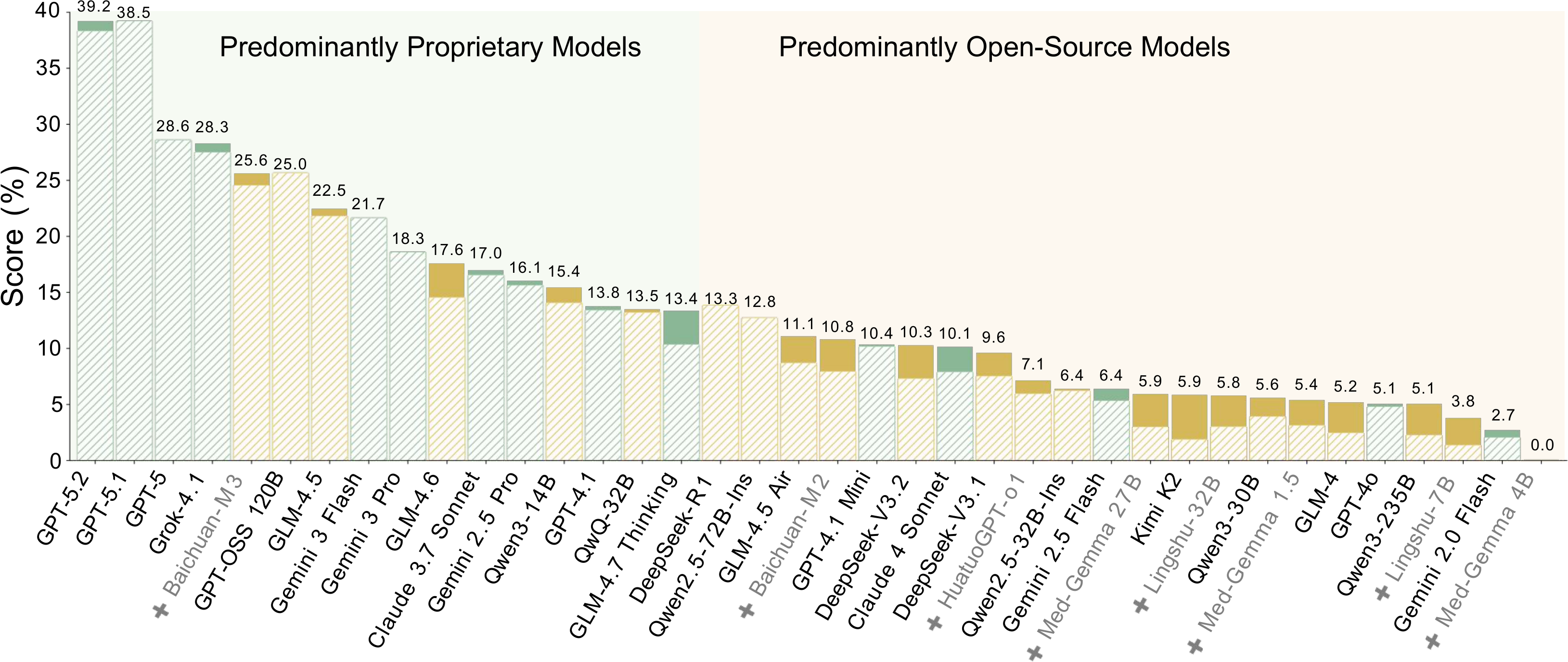}
  \vspace{-2mm}
  \vspace{-2mm}
  \vspace{-2mm}
  \vspace{-2mm}
  \caption{The evaluation results of 38 LLMs on LiveMedBench, categorized into proprietary (green) and open-source (yellow) models. Models marked with a cross (\textbf{+}) are specialized medical models, while others are general-purpose. Solid bars represent performance on the full dataset, while hatched bars indicate performance on cases post-dating the model's knowledge cutoff.}
  \label{fig:model_performance}
  \vspace{-2mm}
  \vspace{-2mm}
\end{figure*}





As illustrated in Figure~\ref{fig:model_performance}, the GPT series maintains a dominant lead on LiveMedBench. GPT-5.2 achieves the state-of-the-art performance with a score of 39.2\%, followed closely by GPT-5.1 at 38.5\%. The Grok-4.1 and Baichuan M3 also demonstrate strong performance, with leading positions among non-GPT models at 28.3\% and 25.6\%, respectively. This hierarchy underscores the continued superiority of highly scaled proprietary architectures in handling complex clinical reasoning tasks.

\textbf{Proprietary vs. Open-Source Models}. A comparative analysis reveals that proprietary models generally outperform open-source alternatives. However, the gap is narrowing significantly. Notably, GPT-OSS 120B (25.0\%) and GLM-4.5 (22.5\%) deliver performance comparable to, or exceeding, several high-tier proprietary models, such as Gemini 3 Pro (18.3\%) and Claude 3.7 Sonnet (17.0\%). Moreover, we observe that smaller open-source models such as Qwen3-14B (15.4\%) and QwQ-32B (13.5\%) demonstrate reasoning capabilities that are competitive with substantially larger proprietary models like Gemini 2.5 Pro (16.1\%). From a deployment perspective, while proprietary models offer peak performance, open-source models offer a viable alternative for privacy-sensitive environments where local deployment is mandatory, and for edge-device deployment where computational resources are limited.

\textbf{General-Purpose vs. Medical-Specific Models}. Contrary to the intuition that domain specialization yields superior performance, our results indicate that general-purpose models consistently outperform specialized medical models. For instance, the general-purpose GPT-5.2 (39.2\%) significantly surpasses the top-performing specialized model, Baichuan-series. This disparity is likely attributable to the ``scaling laws,'' where the massive parameter counts and diverse training of general models provide a more robust foundation than smaller, domain-tuned architectures. Nevertheless, medical models exhibit strong parameter efficiency. For example, Med-Gemma 27B (5.9\%) matches the performance of the general-purpose Gemini 2.5 Flash (6.4\%), suggesting that specialized tuning remains a potent strategy for smaller-scale deployments.

\textbf{Data Contamination and Knowledge Obsolescence}.  To empirically quantify the dual impacts of data contamination and knowledge obsolescence, we evaluated the models on a subset of cases published after their respective knowledge cutoff dates (or release dates, for models with undisclosed cutoffs). As illustrated in Figure~\ref{fig:model_performance}, 84\% of the models (32 out of 38) exhibited a performance degradation when transitioning from the full dataset to the post-cutoff subset. Notably, the GLM series appeared most susceptible to data leakage, with all variants showing a marked decline in accuracy; the most pronounced drop occurred in Kimi-K2, which suffered a 3.99\% decrease. Furthermore, even models without a precipitous drop failed to demonstrate significant performance gains on recent data. Figure~\ref{fig:intro-example} further details the evaluation across different time windows, revealing a consistent downward trend in average performance on contemporary cases (e.g., those post after January 2025). These findings underscore the effect of data contamination, which artificially inflates scores on seen data, and knowledge obsolescence, which leads to failures on unseen medical knowledge. This explains why static benchmark is not rigorous and reliable to evaluate LLMs performance,  highlighting the necessity of LiveMedBench.

\section{Analysis}
We structure our analysis around the following research questions:
\textbf{RQ1.} Is our evaluation pipeline reliable?
\textbf{RQ2.} How does model performance vary across specialties and themes?
\textbf{RQ3.} What are the dominant failure modes?
\textbf{RQ4.} What is the impact of knowledge injection (e.g., retrieval) on performance?
\textbf{RQ5.} How does LiveMedBench compare to existing medical benchmarks?

\subsection{Human Evaluation and Reliability Analysis}


\begin{figure*}[t]
  \centering
  \includegraphics[width=\textwidth]{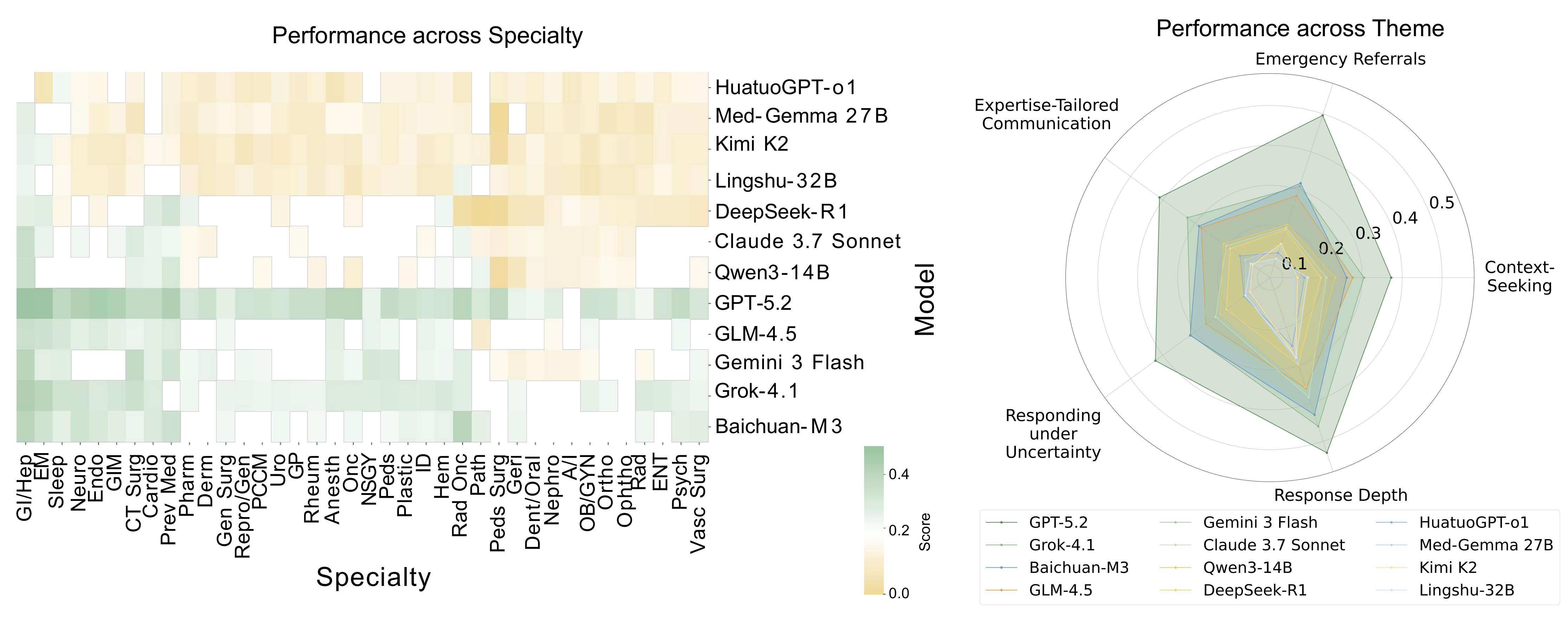}
  \vspace{-2mm}
  \vspace{-2mm}
  \vspace{-2mm}
  \vspace{-2mm}
  \caption{Multi-dimensional performance analysis. (Left) Heatmap illustrating the score distribution of representative models across 38 clinical specialties. (Right) Radar chart depicting the capability profiles of representative models across the five behavioral themes.}
  \label{fig:performance_across_settings}
  \vspace{-2mm}
  \vspace{-2mm}
\end{figure*}


To validate the clinical integrity of LiveMedBench and the reliability of our automated evaluation pipeline, we conducted a human study with two physicians. We randomly select 50 cases and 292 corresponding criterion.

\begin{table}[h]
    \centering
    \caption{Physician assessment of data quality.}
  \vspace{-2mm}
    \label{tab:rater_agreement}
    
    \resizebox{0.75\columnwidth}{!}{%
        \renewcommand{\arraystretch}{1.3}
        
        \begin{tabular}{
            l 
            S[table-format=1.4] 
            S[table-format=1.4] 
            S[table-format=1.4]
        }
            \toprule
            \rowcolor{gray!20}
            \textbf{Task} & 
            {\textbf{Rater 1}} & 
            {\textbf{Rater 2}} & 
            {\textbf{Gwet's AC1}} \\
            \midrule
            
            Narrative \& Query & 0.9795 & 1.0000 & 0.9792 \\
            Physician Advice   & 0.9795 & 0.9800 & 0.9566 \\
            Criterion          & 0.9589 & 0.9110 & 0.8914 \\
            \bottomrule
        \end{tabular}
    }
  \vspace{-2mm}
\end{table}

\textbf{Data Quality and Consensus.} As shown in Table~\ref{tab:rater_agreement}, physicians demonstrated exceptionally high approval rates for our generated data. For the core components of the benchmark, i.e. patient narrative, query and physician advice, the agreement rate (Rater 1 and Rater 2) reached near-perfect levels ($\ge 0.9795$), with an inter-rater reliability (Gwet's AC1) of 0.9792 and 0.9566, respectively. An AC1 score greater than 0.8 indicates ``almost perfect'' agreement, according to~\cite{landis1977}. This confirms that our Multi-Agent Clinical Curation framework successfully synthesizes highly coherent and clinically valid cases. Physicians also validated the generated Rubric Criteria with high consensus (AC1 = 0.8914), indicating that our automated criteria are clinically precise.

\begin{table}[ht]
    \centering
    \caption{Evaluation results comparison. Macro F1 for criteria-level agreement and Pearson Correlation ($\rho$) for case-level alignment with physician scores.}
  \vspace{-2mm}
    \label{tab:eval_comparison}
    
    \resizebox{0.95\columnwidth}{!}{%
        \renewcommand{\arraystretch}{1.3} 
        
        \begin{tabular}{
            l 
            S[table-format=1.2, mode=text] 
            S[table-format=1.2, mode=text] 
            c 
        }
            \toprule
            \rowcolor{gray!20}
            \textbf{Evaluation Method} & 
            {\textbf{Macro F1}} & 
            {\textbf{Correlation $\rho$}} & 
            {\textbf{$P$-value}} \\
            \midrule
            
            Human Inter-Rater              & 0.89 & {--} & {--} \\
            \midrule
            
            LLM-as-a-Judge                 & {--} & 0.26 & 0.07 \\
            \textbf{Rubric-based Grader (Ours)} & 0.76 & 0.54 & \num{6.65e-5} \\
            \bottomrule
        \end{tabular}
    }
  \vspace{-2mm}
\end{table}

\textbf{Reliability of the Rubric-based Grader.} We assessed the alignment between our automated Rubric-based Grader and human experts using the Macro F1 score on a criterion level, following~\cite{healthbench}. As presented in Table~\ref{tab:eval_comparison}, the Human Inter-Rater F1 is 0.89, which serves as the theoretical upper bound for this task. Our Rubric-based Grader achieves a Macro F1 of 0.76, indicating that our grader effectively proxies human judgment.

\textbf{Superiority over LLM-as-a-Judge.} 
We benchmark our method against a standard LLM-as-a-Judge baseline~\cite{zheng2024judging}. Table~\ref{tab:eval_comparison} reveals a critical performance disparity: our Rubric-based Grader achieves a strong Pearson correlation of 0.54 with human scores ($p < 10^{-4}$), whereas the baseline yields a weak, non-significant correlation of 0.26 ($p=0.07$). As shown in Figure~\ref{fig:intro-example}, the baseline exhibits a significant skew towards higher scores, failing to penalize poor responses due to ``length bias'' and a lack of sensitivity to safety hazards. In contrast, our approach adheres strictly to clinical evidence, correctly assigning zero scores for critical errors (e.g., missed contraindications) where the holistic judge failed. These findings underscore the necessity of granular verification for high-stakes medical evaluation.

\subsection{Performance across Specialties and Themes}

To identify specific competency gaps beyond overall scores, we disaggregated model performance across distinct clinical specialties and behavioral themes (Figure~\ref{fig:performance_across_settings}).

\textbf{Performance across Clinical Specialties}. The heat map (Figure~\ref{fig:performance_across_settings}, left) reveals a distinct performance hierarchy: proprietary models (e.g., GPT-5.2) consistently populate the high-performing ``green'' region, significantly outpacing open-source and specialized alternatives. 
Across all models, performance is non-uniform. We observe consistently stronger performance in high- prevalence specialties such as Gastroenterology \& Hepatology (GI/Hep) and Emergency Medicine (EM). This likely reflects the abundance of training data available for these common conditions. Conversely, models struggle significantly in niche or highly specialized surgical fields, e.g. Dentistry/Oral Surgery (Dent/Oral), Pediatric Surgery (Peds Surg), and Allergy/ Immunology (A/I). This suggests a ``knowledge gap'' in current LLMs regarding specialized procedural and diagnostic nuances that are less represented in general training corpora.
We also observed low correlations between general internal medicine specialties and niche fields like Pathology (see Appendix~\ref{app:correlation-analysis}).

\textbf{Performance across Themes}. The radar chart (Figure~\ref{fig:performance_across_settings}, right) corroborates this tiering, with proprietary models occupying the outermost high-performance layers, followed by general open-source models, and finally specialized medical models in the innermost rings. Behaviorally, most models demonstrate strong alignment in \textit{Expertise-Tailored Communication} and \textit{Responding under Uncertainty}. However, a critical safety bottleneck remains in \textit{Context-Seeking}, where even top models frequently fail to proactively gather missing information before providing advice.

\subsection{Error Analysis}

To understand failure modes, we analyzed the top proprietary and open-source models.

\paragraph{Performance by Evaluation Axis.}
As shown in Table~\ref{tab:error_rates_axis}, even SOTA models exhibit error rates exceeding 50\% on most axes. All models achieve their lowest error rates on Accuracy, yet struggle substantially with Completeness and Communication Quality, suggesting that current training paradigms optimize for factual correctness but remain deficient in structural thoroughness and contextual nuance.

\begin{table}[t]
    \centering
    \caption{Error rates of representative LLMs. Worst rates (highest errors) are \textbf{bolded}.}
  \vspace{-2mm}
    \label{tab:error_rates_axis}
    
    \resizebox{\linewidth}{!}{
        \renewcommand{\arraystretch}{1.3}
        \setlength{\tabcolsep}{4pt} 
        
        \begin{tabular}{
            l 
            S[table-format=2.2] 
            S[table-format=2.2] 
            S[table-format=2.2] 
            S[table-format=2.2] 
            S[table-format=2.2] 
            S[table-format=2.2, mode=text] 
        }
            \toprule
            \rowcolor{gray!20}
            \textbf{Metric} & 
            {\textbf{\makecell{Baichuan\\M3}}} & 
            {\textbf{\makecell{GPT\\5.2}}} & 
            {\textbf{\makecell{Grok\\4.1}}} & 
            {\textbf{\makecell{GPT-\\OSS}}} & 
            {\textbf{\makecell{GLM-\\4.5}}} & 
            {\textbf{\makecell{Qwen3-\\14B}}} \\
            \midrule
            \rowcolors{2}{gray!6}{white}
            Accuracy          & 43.98 & 37.54 & 42.01 & 45.46 & 45.84 & \textbf{50.26} \\
            Completeness      & 64.00 & 49.59 & 58.41 & 55.23 & 62.26 & \textbf{68.46} \\
            Comm. Quality     & \textbf{66.97} & 50.37 & 56.89 & 57.21 & 60.61 & 66.96 \\
            Context Awareness & 59.62 & 48.33 & 55.85 & 56.40 & 58.09 & \textbf{65.13} \\
            Safety            & 48.36 & 38.20 & 44.88 & 44.97 & 47.42 & \textbf{52.93} \\
            \bottomrule
        \end{tabular}
    }
  \vspace{-2mm}
\end{table}

\paragraph{Qualitative Root Causes.}
We further audited the bottom-100 scoring cases per model using GPT-4.1 to classify root causes into seven error types (method and results in Appendix~\ref{app:error-analysis}). Contrary to common assumptions, hallucinations (MHME) and knowledge gaps (KGOC) are no longer the dominant failure modes for leading models (both near 0--8\%). Instead, the vast majority of errors stem from \textit{Contextual Neglect and Integration Failure} (CNIF: 35--48\%) and \textit{Guideline Overgeneralization} (GOPR: 22--32\%), indicating that models possess the requisite medical facts but struggle to tailor them to patient-specific constraints such as comorbidities or contraindications. Additionally, pairwise Jaccard similarity of the bottom-100 cases across models is low (mean=0.24, Figure~\ref{fig:case_correlation}), confirming that failures reflect model-specific capability gaps rather than data quality artifacts.

\subsection{Impact of Knowledge Injection}


Here we investigate whether the performance degradation on unseen data stems from inherent reasoning deficits or knowledge obsolescence. We isolated a subset of ``fresh'' cases from January 2026 and benchmark performance across three settings: (1) \textbf{Baseline}: Model performance on full dataset under default settings; (2) \textbf{Closed-Book}: Standard inference on the subset; and (3) \textbf{Open-Book}: Inference on the Jan 2026 subset augmented with knowledge retrieval ~\cite{medrag} to provide access to external medical knowledge.

\begin{table}[ht]
    \centering
    \caption{Closed book performance versus open book performance (\%).}
    \label{tab:knowledge-injection}

    \resizebox{\columnwidth}{!}{%
        \renewcommand{\arraystretch}{1.3} 
        
        \begin{tabular}{
            l 
            S[table-format=2.2, detect-weight, mode=text] 
            S[table-format=2.2, detect-weight, mode=text] 
            S[table-format=2.2, detect-weight, mode=text] 
            S[table-format=2.2, detect-weight, mode=text] 
            S[table-format=2.2, detect-weight, mode=text] 
            S[table-format=2.2, detect-weight, mode=text]
        }
            \toprule
            \rowcolor{gray!20}
            \textbf{Mode} & 
            {\textbf{\makecell{Baichuan\\M3}}} & 
            {\textbf{\makecell{GPT\\5.2}}} & 
            {\textbf{\makecell{Grok\\4.1}}} & 
            {\textbf{\makecell{GPT-OSS\\120B}}} & 
            {\textbf{\makecell{GLM\\4.5}}} & 
            {\textbf{\makecell{Qwen3\\14B}}} \\
            \midrule
            
            Baseline    & 25.61 & 39.23 & 28.28 & 25.03 & 22.46 & 15.45 \\
            \midrule
            
            Closed-Book & 24.54 & 36.84 & 24.21 & 23.24 & 16.19 & 8.68 \\
            Open-Book   & \textbf{25.21} & \textbf{37.17} & \textbf{27.39} & \textbf{24.18} & \textbf{20.45} & \textbf{15.47} \\
            \bottomrule
        \end{tabular}
    }
\end{table}

As shown in Table~\ref{tab:knowledge-injection}, enabling access to external knowledge yields universal performance improvements across all model architectures. This suggests that failures on new cases are partially driven by \textit{retrieval failures}  rather than \textit{logic failures}. Furthermore, the significant drop between the ``Baseline'' and ``Closed-Book'' scores quantifies the distinct impact of data contamination and knowledge obsolescence on static LLMs.


\begin{table}[ht]
    \centering
    \caption{Medical Benchmark Comparison.}
  \vspace{-2mm}
    \label{tab:med-benchmarks}
    
    \resizebox{0.86\columnwidth}{!}{%
        \renewcommand{\arraystretch}{1.2}
        \setlength{\tabcolsep}{6pt}
        
        \begin{tabular}{l cccc}
            \toprule
            \textbf{Benchmark} & 
            \textbf{\makecell{Free\\Response}} & 
            \textbf{\makecell{Multi-\\lingual}} & 
            \textbf{\makecell{Case-specific\\Rubric}} & 
            \textbf{\makecell{Case\\Update}} \\
            \midrule
            MedQA (2021)         & \redx & \greencheck & \redx & \redx \\
            MultiMedQA (2023)    & \greencheck & \redx & \redx & \redx \\
            MedBench (2024)      & \redx & \redx & \redx & \redx \\
            CMB (2024)           & \greencheck & \redx & \redx & \redx \\
            CliMedBench (2024)   & \greencheck & \redx & \redx & \redx \\
            MedJourney (2024)    & \greencheck & \greencheck & \redx & \redx \\
            GMAI-MMBench (2024)  & \redx & \redx & \redx & \redx \\
            MedXpertQA (2025)    & \redx & \redx & \redx & \redx \\
            DyReMe (2025)        & \greencheck & \redx & \redx & \redx \\
            MedPerturb (2025)    & \greencheck & \redx & \redx & \redx \\
            HealthBench (2025)   & \greencheck & \redx & \greencheck & \redx \\
            \midrule
            \rowcolor{gray!15}
            \textbf{LiveMedBench (Ours)} & \greencheck & \greencheck & \greencheck & \greencheck \\
            \bottomrule
        \end{tabular}%
    }
  \vspace{-2mm}
\end{table}

\subsection{Medical Benchmark Comparison}



To contextualize the contribution of LiveMedBench within the broader landscape of medical LLM evaluation, we conducted both a qualitative feature comparison and a quantitative difficulty assessment against established benchmarks.

As summarized in Table~\ref{tab:med-benchmarks}, existing benchmarks typically suffer from one or more limitations. LiveMedBench distinguishes itself as the only benchmark that simultaneously integrates (1) Free-response generation, (2) Multi-lingual support, (3) Granular case-specific rubrics, and (4) Live case updates. This unique combination addresses the critical need for an evaluation framework that is both clinically rigorous and temporally adaptive.

\begin{table}[ht]
    \centering
    \caption{Comparison of ours and HealthBench.}
  \vspace{-2mm}
    \label{tab:healthbench_vs_livemedbench}
    
    \resizebox{0.88\columnwidth}{!}{%
        \renewcommand{\arraystretch}{1.3}
        
        \begin{tabular}{
            l 
            S[table-format=1.4, detect-weight, mode=text] 
            S[table-format=1.4, detect-weight, mode=text] 
            S[table-format=1.4, detect-weight, mode=text] 
            S[table-format=1.4, detect-weight, mode=text]
        }
            \toprule
            \rowcolor{gray!20}
            \textbf{Benchmark} & 
            {\textbf{\makecell{Gemini\\2.5 Pro}}} & 
            {\textbf{\makecell{GPT-4.1}}} & 
            {\textbf{\makecell{Claude 3.7\\Sonnet}}} & 
            {\textbf{\makecell{GPT-4o}}} \\
            \midrule
            
            HealthBench      & \textbf{0.52}   & 0.48   & 0.35   & 0.32   \\
            HealthBench Hard & \textbf{0.19}   & 0.16   & 0.02   & 0      \\
            \midrule
            
            \textbf{LiveMedBench} & 0.1606 & 0.1379 & \textbf{0.1699} & 0.0506 \\
            \bottomrule
        \end{tabular}
    }
  \vspace{-2mm}
\end{table}

To assess the challenge level of our dataset, we benchmarked four LLMs (Gemini 2.5 Pro, GPT-4.1, Claude 3.7 Sonnet, GPT-4o) on LiveMedBench and compared the results against HealthBench, the closest competitor. As shown in Table~\ref{tab:healthbench_vs_livemedbench}, models exhibit significantly lower scores on LiveMedBench compared to the standard HealthBench dataset. Notably, the performance on LiveMedBench aligns with the ``HealthBench Hard'' subset (e.g., Gemini 2.5 Pro scores 0.19 on Hard vs. 0.16 on ours). This performance gap underscores the complexity of our dataset, leaving significant room for future model improvement.

\section{Conclusion}

We introduced LiveMedBench, a continually updated, contamination free medical benchmark for evaluating LLMs. Our Multi-Agent Clinical Curation Framework transforms noisy online discussions into evidence-backed clinical cases, and our Automated Rubric-based Evaluation Framework enables granular, objective assessment in alignment with physician experts. Evaluation of 38 LLMs reveals consistent performance degradation on post-cutoff cases, validating the necessity of live benchmarking.


\bibliographystyle{ACM-Reference-Format}
\bibliography{livemedbench}

\newpage
\appendix

\section{Dataset}
\label{app:dataset}

\subsection{Online Medical Communities}
\label{app:platforms}

\subsubsection{DXY}
\textbf{DXY}\footnote{\url{https://dxy.com/}} is widely recognized as the largest online academic community for healthcare professionals in China. It features a rigorous Question-and-Answer (Q\&A) module where inquiries are addressed by verified licensed physicians. The platform covers a comprehensive range of clinical specialties and is distinguished by its strict verification process, ensuring that the medical advice provided is professional, reliable, and grounded in clinical practice.

\subsubsection{MedLive}
\textbf{MedLive}\footnote{\url{https://www.medlive.cn/}} is a premier professional medical platform in China, primarily designed to support clinical decision-making and academic exchange among physicians. Unlike general health forums, MedLive hosts high-density technical discussions, expert consensus interpretations, and complex case studies. Its data is characterized by a high degree of professional terminology and clinical depth, making it an invaluable source for evaluating an LLM's ability to handle specialized medical contexts.

\subsubsection{iCliniq}
\textbf{iCliniq}\footnote{\url{https://www.icliniq.com/qa/}} is a leading global telemedicine platform that connects patients with doctors across more than 80 medical specialties. The platform operates on a "second opinion" model, where users post medical queries that are answered by credentialed and verified medical experts. The interactions on iCliniq typically follow a structured format comprising a detailed patient query followed by a comprehensive, empathetic, and evidence-based response from a qualified physician, providing high-quality English-language training and evaluation data.

\subsubsection{SDN}
\textbf{The Student Doctor Network (SDN)}\footnote{\url{https://forums.studentdoctor.net/forums/}} is one of the largest non-profit educational communities for healthcare students and professionals in North America. While it serves a broad educational purpose, its specific clinical forums facilitate peer-to-peer discussions among medical students, residents, and attending physicians. These dialogues often involve collaborative clinical reasoning, detailed case breakdowns, and debates on treatment protocols, offering a unique "clinician-to-clinician" perspective distinct from standard doctor-patient interactions.

\subsection{License}
\label{app:license}

Similar to~\cite{livecodebench}, we scrape only the content of post discussions from medical websites– DXY, MedLive, iCliniq and SDN. Further, we
only scrape publicly visible portions of websites, avoiding any data collection that might be pay walled or require login or interaction with the website. Following~\cite{livecodebench} we abide by Fair Use §107: ``the fair use of a copyrighted work, including such use by ... scholarship, or research, is not an infringement of copyright'', where fair use is determined by ``the purpose and
character of the use, including whether such use is of a commercial nature or is for nonprofit educa tional purposes'', ``the amount and substantiality of the portion used in relation to the copyrighted work as a whole'', and ``the effect of the use upon the potential market for or value of the copy righted work.'' Finally, we use the collected problems for academic purposes only and in addition,
do not train on the collected problems.

\subsection{Theme and Axis Definitions}
\label{app:themes-axes}

LiveMedBench adopts and adapts the evaluation taxonomy introduced by HealthBench~\cite{healthbench}. We retain five of HealthBench's seven \textbf{Themes} (dropping \textit{Global Health} and \textit{Health Data Tasks}, which are less represented in the online medical communities we target) and replace the \textit{Instruction Following} axis with \textit{Safety} to better capture contraindication and harm-prevention failures in open-ended clinical advice.

\subsubsection{Themes}
Themes are high-level categories of health-related tasks that reflect distinct challenges in real-world clinical interactions. Each case is assigned exactly one theme.

\begin{itemize}
    \item \textbf{Emergency Referrals.} Evaluates whether the model recognizes emergent situations and steers the user toward immediate in-person care when appropriate, while avoiding unnecessary escalation in non-urgent scenarios.
    \item \textbf{Context-Seeking.} Assesses whether the model identifies when key clinical context is missing and proactively requests the most informative additional information before providing a definitive answer.
    \item \textbf{Expertise-Tailored Communication.} Tests whether the model infers the user's level of medical expertise (e.g., layperson vs.\ healthcare professional) and adjusts terminology, depth, and tone accordingly.
    \item \textbf{Responding under Uncertainty.} Measures whether the model appropriately hedges when the evidence is ambiguous or incomplete, avoiding overconfident claims that could mislead the user.
    \item \textbf{Response Depth.} Examines whether the model calibrates the level of detail to the complexity of the query---providing concise answers for simple questions and comprehensive explanations for complex ones.
\end{itemize}

\subsubsection{Axes}
Axes define the behavioral dimensions along which each rubric criterion evaluates a model response. A single case may contain criteria spanning multiple axes.

\begin{itemize}
    \item \textbf{Accuracy.} Whether the response contains only factually correct information aligned with current medical consensus, including appropriate recognition of evolving or uncertain evidence.
    \item \textbf{Completeness.} Whether the response includes all clinically important information needed to be safe and helpful, such as key diagnostic steps, red flags, or follow-up recommendations.
    \item \textbf{Communication Quality.} Whether the response is well-structured, clear, and uses a level of technical depth and vocabulary matched to the user.
    \item \textbf{Context Awareness.} Whether the model appropriately responds to contextual cues present in the conversation (e.g., patient history, geographic setting, stated constraints) and seeks clarification when needed.
    \item \textbf{Safety.} Whether the model avoids recommending contraindicated treatments, recognizes potential harms, and includes necessary warnings or disclaimers for high-risk situations.
\end{itemize}

\subsection{Details of Data Statistics}
\label{app:data-statistics}

LiveMedBench currently comprises a total of 2,756 clinical cases and 16,702 unique rubric criteria, establishing a robust foundation for granular medical evaluation. The detailed distribution of the dataset is illustrated in Figure~\ref{fig:statistics}.

\paragraph{Clinical Specialties.} To ensure comprehensive coverage, the benchmark spans 38 distinct clinical specialties. As shown in Figure~\ref{fig:statistics}(a), the distribution is naturally long-tailed, reflecting real-world prevalence. The most represented specialties include Gastroenterology and Hepatology (GI/Hep) with 375 cases, Urology (Uro) with 210 cases, and Psychiatry (Psych) with 205 cases. This diversity ensures that models are tested across a wide spectrum of medical knowledge, from general practice to highly specialized fields like Neuro-Oncology.

\paragraph{Linguistic Diversity.} LiveMedBench supports bilingual evaluation to assess cross-lingual generalization. As depicted in Figure~\ref{fig:statistics}(c), the dataset includes 1,568 English cases (sourced from iCliniq and SDN) and 1,188 Chinese cases (sourced from DXY and Medlive), maintaining a balanced representation to rigorously evaluate performance in both linguistic contexts.

\paragraph{Health Contexts.} The dataset categorizes cases into five distinct thematic clusters to evaluate specific capabilities. As shown in Figure~\ref{fig:statistics}(b), Expertise-Tailored Communication is the dominant theme, accounting for 47.4\% of the cases, followed by Responding under Uncertainty (27.2\%) and Response Depth (17.6\%). Smaller proportions are dedicated to Context-Seeking (4.5\%) and Emergency Referrals (3.3\%), focusing on critical safety and interactive diagnosis skills.

\paragraph{Behavioral Dimensions.} Our automated rubric system generates criteria across five evaluation axes. As illustrated in Figure~\ref{fig:statistics}(d), the majority of criteria focus on Accuracy (51.2\%), ensuring that clinical correctness remains the primary metric. Completeness accounts for 24.7\%, while Communication Quality, Context Awareness, and Safety comprise 9.4\%, 7.8\%, and 6.9\% respectively. This multi-dimensional structure allows for a holistic assessment of model performance beyond mere factual recall.

\paragraph{Granularity of Evaluation.} A key feature of LiveMedBench is its case-specific evaluation. Figure~\ref{fig:statistics}(e) shows the distribution of criteria count per case, which follows a normal-like distribution. Cases are evaluated on anywhere from a minimum of 2 to a maximum of 19 criteria, with an average of 6.06 criteria per case. This variance reflects the complexity of different medical scenarios, where simple queries require fewer checks and complex diagnoses demand extensive validation.

\section{Experimental Setup}
\label{app:model-specs}

\subsection{Models}
\label{app:models}

Details of models considered in our study is in Table~\ref{tab:model_details}.

\begin{table*}[t]
\centering
\caption{Details of models considered in our study.}
\label{tab:model_details}
\scriptsize
\setlength{\tabcolsep}{4pt}
\resizebox{0.6\textwidth}{!}{%
\begin{tabular}{@{}llll@{}}
\toprule
\textbf{Model Name} & \textbf{Model ID} & \textbf{Date} & \textbf{Link} \\
\midrule
GPT-5.2 & gpt\_5.2 & 2025/08 & \href{https://platform.openai.com/docs/models/gpt-5.2}{openai.com} \\
Baichuan-M2 & baichuan\_m2 & 2025/08 & \href{https://huggingface.co/collections/baichuan-inc/baichuan-m2/}{huggingface.co} \\
GPT-5.1 & gpt\_5\_1 & 2024/09 & \href{https://platform.openai.com/docs/models/gpt-5.1}{openai.com} \\
GPT-4.1 Mini & gpt\_4\_1\_mini & 2024/06 & \href{https://platform.openai.com/docs/models/gpt-4.1-mini}{openai.com} \\
GPT-5 & gpt\_5 & 2024/10 & \href{https://platform.openai.com/docs/models/gpt-5}{openai.com} \\
DeepSeek-V3.2 & deepseek\_v3.2 & 2025/10 & \href{https://huggingface.co/deepseek-ai/DeepSeek-V3.2-Exp}{huggingface.co} \\
Grok-4.1 & grok\_4\_1 & 2024/11 & \href{https://x.ai/news/grok-4-1/}{x.ai} \\
Claude 4 Sonnet & claude-sonnet-4 & 2025/03 & \href{https://www.anthropic.com/news/claude-4}{anthropic.com} \\
Baichuan-M3 & baichuan\_m3\_plus & 2026/01 & \href{https://huggingface.co/collections/baichuan-inc/baichuan-m3/}{huggingface.co} \\
DeepSeek-V3.1 & deepseek\_v3\_1 & 2025/08 & \href{https://huggingface.co/deepseek-ai/DeepSeek-V3.1}{huggingface.co} \\
GPT-OSS 120B & gpt\_oss\_120b & 2024/06 & \href{https://huggingface.co/openai/gpt-oss-120b}{huggingface.co} \\
HuatuoGPT-o1 & huatuogpt\_o1\_8b & 2024/12 & \href{https://huggingface.co/FreedomIntelligence/HuatuoGPT-o1-8B}{huggingface.co} \\
GLM-4.5 & glm\_4\_5 & 2025/08 & \href{https://huggingface.co/zai-org/GLM-4.5/}{huggingface.co} \\
Qwen2.5-32B & qwen2\_5\_32b\_instruct & 2023/12 & \href{https://huggingface.co/Qwen/Qwen2.5-32B-Instruct}{huggingface.co} \\
Gemini 3 Flash & gemini\_3\_flash & 2025/01 & \href{https://blog.google/products-and-platforms/products/gemini/gemini-3-flash/}{gemini.google.com} \\
Gemini 2.5 Flash & gemini\_2\_5\_flash & 2025/01 & \href{https://deepmind.google/technologies/gemini/}{gemini.google.com} \\
Gemini 3 Pro & gemini\_3\_pro & 2025/01 & \href{https://blog.google/products-and-platforms/products/gemini/gemini-3/}{gemini.google.com} \\
Med-Gemma 27B & medgemma\_27b & 2025/05 & \href{https://huggingface.co/google/medgemma-27b-text-it}{huggingface.co} \\
GLM-4.6 Think & glm\_4\_6\_think & 2025/10 & \href{https://huggingface.co/collections/zai-org/glm-46/}{huggingface.co} \\
Kimi K2 & kimi\_k2 & 2025/07 & \href{https://huggingface.co/collections/moonshotai/kimi-k2/}{huggingface.co} \\
Claude 3.7 & claude-3-7-sonnet & 2024/10 & \href{https://www.anthropic.com/news/claude-3-7-sonnet}{anthropic.com} \\
Lingshu-32B & lingshu\_32b & 2025/06 & \href{https://huggingface.co/collections/lingshu-medical-mllm/lingshu-mllms/}{huggingface.co} \\
Gemini 2.5 Pro & gemini\_2\_5\_pro & 2025/01 & \href{https://blog.google/innovation-and-ai/models-and-research/google-deepmind/gemini-model-thinking-updates-march-2025/}{gemini.google.com} \\
Qwen3-30B & qwen3\_30b\_a3b & 2025/07 & \href{https://huggingface.co/Qwen/Qwen3-30B-A3B}{huggingface.co} \\
Qwen3-14B & qwen3\_14b & 2025/05 & \href{https://huggingface.co/Qwen/Qwen3-14B}{huggingface.co} \\
Med-Gemma 1.5 & medgemma\_1.5 & 2026/01 & \href{https://huggingface.co/google/medgemma-1.5-4b-it}{huggingface.co} \\
GPT-4.1 & gpt\_4\_1 & 2024/06 & \href{https://platform.openai.com/docs/models/gpt-4.1}{openai.com} \\
GLM-4 & glm\_4 & 2025/04 & \href{https://huggingface.co/collections/zai-org/glm-4/}{huggingface.co} \\
QwQ-32B & qwq\_32b & 2024/11 & \href{https://huggingface.co/Qwen/QwQ-32B}{huggingface.co} \\
GPT-4o & gpt\_4o & 2023/10 & \href{https://platform.openai.com/docs/models/gpt-4o}{openai.com} \\
GLM-4.7 & glm\_4\_7 & 2025/12 & \href{https://huggingface.co/collections/zai-org/glm-47/}{huggingface.co} \\
Qwen3-235B & qwen3\_235b\_a22b & 2025/07 & \href{https://huggingface.co/Qwen/Qwen3-235B-A22B}{huggingface.co} \\
DeepSeek-R1 & deepseek\_r1 & 2025/01 & \href{https://huggingface.co/deepseek-ai/DeepSeek-R1}{huggingface.co} \\
Lingshu-7B & lingshu\_7b & 2025/06 & \href{https://huggingface.co/collections/lingshu-medical-mllm/lingshu-mllms/}{huggingface.co} \\
Qwen2.5-72B & qwen2\_5\_72b\_instruct & 2023/12 & \href{https://huggingface.co/Qwen/Qwen2.5-72B-Instruct}{huggingface.co} \\
Gemini 2.0 Flash & gemini\_2\_0\_flash & 2024/08 & \href{https://blog.google/innovation-and-ai/models-and-research/google-deepmind/google-gemini-ai-update-december-2024/}{gemini.google.com} \\
GLM-4.5 Air & glm\_4\_5\_air & 2025/08 & \href{https://huggingface.co/collections/zai-org/glm-45/}{huggingface.co} \\
Med-Gemma 4B & medgemma\_4b & 2025/05 & \href{https://huggingface.co/google/medgemma-4b-it}{huggingface.co} \\
\bottomrule
\end{tabular}%
}
\end{table*}

\subsection{Details in Curation Framework}
\label{app:curation-details}

Prompts for the screener are in Figure~\ref{fig:screener_L_N}, Figure~\ref{fig:screener_Q} and Figure~\ref{fig:screener_L_A}. 

Prompts for the validator are in Figure~\ref{fig:validator_cc}, Figure~\ref{fig:validator_inf} and Figure~\ref{fig:validator_align}. 

The prompt for the comtroller is in Figure~\ref{fig:controller}.

\begin{figure*}[t]
    \centering
    \begin{tcolorbox}[
        colback=white,          
        colframe=black!75,      
        title=\textbf{System Prompt: the Screener}, 
        fonttitle=\bfseries\sffamily,
        boxrule=0.8pt,          
        arc=2mm,                
        width=\textwidth        
    ]
    \small\ttfamily 
\textbf{Role:} You are an expert Medical Data Structurer.

\textbf{Task:} Your goal is to extract the \textbf{Patient Narrative ($L_N$)} from the provided raw dialogue. This corresponds to the \textbf{Subjective} and \textbf{Objective} sections of the SOAP note.

\textbf{Instructions:}
\begin{itemize}
    \item \textbf{Decompose:} Analyze the patient's text and classify information into Subjective (reported symptoms/history) and Objective (measurable data/facts).
    \item \textbf{Filter:} Remove all non-clinical text, including greetings, emotional fluff, gratitude, and irrelevant conversational fillers.
    \item \textbf{Subjective Extraction:} Focus on the History of Present Illness (HPI), chief complaints, pain levels, and description of symptoms.
    \item \textbf{Objective Extraction:} Focus on specific data points such as current medications, dosage, vital signs, or lab results.
    \item \textbf{Anonymity:} Ensure no personally identifiable information (names, specific locations) is included.
\end{itemize}

\textbf{Input:} [Raw Patient Dialogue]

\textbf{Output Format:}
\begin{itemize}
    \item \textbf{Subjective:} [List of patient's history, chief complaints, and symptoms]
    \item \textbf{Objective:} [List of verifiable data: vitals, lab results, and medications (or "None" if absent)]
\end{itemize}
    \end{tcolorbox}
    \caption{The prompt template used for the screener to extract L\_N.}
    \label{fig:screener_L_N}
\end{figure*}

\begin{figure*}[t]
    \centering
    \begin{tcolorbox}[
        colback=white,          
        colframe=black!75,      
        title=\textbf{System Prompt: the Screener}, 
        fonttitle=\bfseries\sffamily,
        boxrule=0.8pt,          
        arc=2mm,                
        width=\textwidth        
    ]
    \small\ttfamily 
\textbf{Role:} You are a Medical Intent Analyzer.

\textbf{Task:} Identify and refine the \textbf{Primary Query ($Q$)} from the patient's submission.

\textbf{Instructions:}
\begin{itemize}
    \item \textbf{Identify:} pinpoint the core medical question the user is asking.
    \item \textbf{Explicitize:} If the query is implicit (e.g., "I am worried about this lump"), convert it into an explicit clinical question (e.g., "What are the potential causes of this lump and does it require urgent care?").
    \item \textbf{Contextualize:} Ensure the query is self-contained and understandable without reading the full narrative.
\end{itemize}

\textbf{Input:} [Raw Patient Dialogue]

\textbf{Output Format:} A single, clear question string.
    \end{tcolorbox}
    \caption{The prompt template used for the screener to extract Q.}
    \label{fig:screener_Q}
\end{figure*}

\begin{figure*}[t]
    \centering
    \begin{tcolorbox}[
        colback=white,          
        colframe=black!75,      
        title=\textbf{System Prompt: the Screener}, 
        fonttitle=\bfseries\sffamily,
        boxrule=0.8pt,          
        arc=2mm,                
        width=\textwidth        
    ]
    \small\ttfamily 
\textbf{Role:} You are a Senior Clinical Consultant.

\textbf{Task:} Extract the \textbf{Physician Advice ($L_A$)} from the verified doctor's response. This corresponds to the \textbf{Assessment} and \textbf{Plan} sections of the SOAP note.

\textbf{Instructions:}
\begin{itemize}
    \item \textbf{Extract:} Identify the doctor's primary diagnosis (or differential diagnosis), reasoning, and specific recommendations.
    \item \textbf{Actionability:} Focus on actionable steps: medications, lifestyle changes, "red flag" warnings, or referral instructions (e.g., "See a cardiologist").
    \item \textbf{Factuality:} Strictly adhere to the doctor's original text. Do not hallucinate or add medical advice not present in the source.
\end{itemize}

\textbf{Input:} [Verified Physician Response]

\textbf{Output Format:}
\begin{itemize}
    \item \textbf{Assessment:} [List of doctor's reasoning/diagnosis]
    \item \textbf{Plan:} [List of actionable recommendations]
\end{itemize}
    \end{tcolorbox}
    \caption{The prompt template used for the screener to extract L\_A.}
    \label{fig:screener_L_A}
\end{figure*}

\begin{figure*}[t]
    \centering
    \begin{tcolorbox}[
        colback=white,          
        colframe=black!75,      
        title=\textbf{System Prompt: the Validator}, 
        fonttitle=\bfseries\sffamily,
        boxrule=0.8pt,          
        arc=2mm,                
        width=\textwidth        
    ]
    \small\ttfamily 
\textbf{Role:} You are a Medical Taxonomy Expert based on Ely's Taxonomy of Clinical Questions.

\textbf{Task:} Evaluate the \textbf{Primary Query ($Q$)} to determine if it constitutes a clinically meaningful Chief Complaint (CC) or medical request.

\textbf{Instructions:}
\begin{itemize}
    \item \textbf{Analyze:} Does $Q$ fall into a valid clinical category (e.g., Symptom, Diagnosis, Management, Prognosis, Therapy)?
    \item \textbf{Filter:} Reject queries that are:
    \begin{itemize}
        \item Nonsense or gibberish.
        \item Purely social/chit-chat without medical intent.
        \item Asking for illegal acts (e.g., manufacturing drugs).
        \item Too vague to be answered (e.g., "I feel bad").
    \end{itemize}
    \item \textbf{Output:} Return a binary score (1 for Valid, 0 for Invalid).
\end{itemize}

\textbf{Input:} [User Query $Q$]

\textbf{Output Format (JSON):}
\begin{verbatim}
{
  "reasoning": "Brief explanation...",
  "score": 1  // or 0
}
\end{verbatim}
    \end{tcolorbox}
    \caption{The prompt template used for the validator to compute S\_cc.}
    \label{fig:validator_cc}
\end{figure*}

\begin{figure*}[t]
    \centering
    \begin{tcolorbox}[
        colback=white,          
        colframe=black!75,      
        title=\textbf{System Prompt: the Validator}, 
        fonttitle=\bfseries\sffamily,
        boxrule=0.8pt,          
        arc=2mm,                
        width=\textwidth        
    ]
    \small\ttfamily 
\textbf{Role:} You are a Clinical Auditor.

\textbf{Task:} Calculate the \textbf{Narrative Sufficiency Score ($S_{inf}$)}. You must determine if the Patient Narrative ($L_N$) provides enough context to safely answer the Query ($Q$).

\textbf{Instructions:}
\begin{enumerate}
    \item \textbf{Step 1 (Derive Requirements):} List the atomic clinical facts ($S_{req}$) absolutely necessary to answer $Q$ (e.g., for "Headache", you need: Onset, Severity, Frequency, Triggers).
    \item \textbf{Step 2 (Verify):} Check $L_N$. For each item in $S_{req}$, determine if it is present/entailed in the narrative.
    \item \textbf{Step 3 (Score):} Calculate the ratio of present items to required items.
\end{enumerate}

\textbf{Input:}
\begin{itemize}
    \item \textbf{Query ($Q$):} ...
    \item \textbf{Narrative ($L_N$):} ...
\end{itemize}

\textbf{Output Format (JSON):}
\begin{verbatim}
{
  "required_items": ["item1", "item2", "item3", ...],
  "present_items": ["item1", "item3"],
  "score": 0.66  // float between 0.0 and 1.0
}
\end{verbatim}
    \end{tcolorbox}
    \caption{The prompt template used for the validator to compute S\_inf.}
    \label{fig:validator_inf}
\end{figure*}

\begin{figure*}[t]
    \centering
    \begin{tcolorbox}[
        colback=white,          
        colframe=black!75,      
        title=\textbf{System Prompt: the Validator}, 
        fonttitle=\bfseries\sffamily,
        boxrule=0.8pt,          
        arc=2mm,                
        width=\textwidth        
    ]
    \small\ttfamily 
\textbf{Role:} You are a Senior Medical Guideline Validator.

\textbf{Task:} Verify the \textbf{Physician Advice ($L_A$)} against the provided \textbf{Retrieved Evidence ($E$)}.

\textbf{Instructions:}
\begin{itemize}
    \item \textbf{Verify:} Compare each claim $a_j$ in $L_A$ against the retrieved evidence text.
    \item \textbf{Score each claim ($v(a_j)$):}
    \begin{itemize}
        \item \textbf{1.0 (Supported):} Explicitly confirmed by the evidence.
        \item \textbf{0.5 (Neutral):} Not mentioned in evidence, but clinically standard/safe (common sense).
        \item \textbf{0.0 (Contradicted):} Directly contradicts the evidence (Hallucination/Safety Risk).
    \end{itemize}
\end{itemize}

\textbf{Input:}
\begin{itemize}
    \item \textbf{Physician Advice ($L_A$):} ...
    \item \textbf{Retrieved Guidelines ($E$):} ...
\end{itemize}

\textbf{Output Format (JSON):}
\begin{verbatim}
{
  "claims_analysis": [
    {"claim": "Take 500mg Tylenol", "verification": "Supported by text...", "score": 1.0},
    {"claim": "Run a marathon immediately", "verification": "Contradicts rest protocols", "score": 0.0}
  ],
  ...
}
\end{verbatim}
    \end{tcolorbox}
    \caption{The prompt template used for the validator to compute S\_align.}
    \label{fig:validator_align}
\end{figure*}

\begin{figure*}[t]
    \centering
    \begin{tcolorbox}[
        colback=white,          
        colframe=black!75,      
        title=\textbf{System Prompt: the Controller}, 
        fonttitle=\bfseries\sffamily,
        boxrule=0.8pt,          
        arc=2mm,                
        width=\textwidth        
    ]
    \small\ttfamily 
\textbf{Role:} You are a Medical Quality Assurance Specialist and Editor.

\textbf{Task:} You have two sequential objectives:
\begin{enumerate}
    \item \textbf{Audit (Fabrication Detection):} Verify that every fact in the \textit{Structured Narrative} ($L_N$) and \textit{Structured Advice} ($L_A$) is explicitly supported by the \textit{Original Raw Thread} ($T$).
    \item \textbf{Synthesis:} Rewrite the structured points into coherent natural language to form the final dataset entry.
\end{enumerate}

\textbf{Instructions:}
\begin{itemize}
    \item \textbf{Step 1: The Veto Check:}
    \begin{itemize}
        \item Compare $L_N$ and $L_A$ against the Raw Thread $T$.
        \item If $L_N$ or $L_A$ contains specific details (e.g., drug names, dosages, exact dates, lab values) that are NOT present in $T$, you must \textbf{REJECT} the case immediately.
        \item \textit{Reasoning:} We cannot allow the AI to "hallucinate" details to make the story better.
    \end{itemize}
    
    \item \textbf{Step 2: Natural Language Synthesis:}
    \begin{itemize}
        \item \textbf{Generate $N$ (Narrative):} Convert the structured patient facts ($L_N$) into a fluent paragraph.
        \item \textbf{Generate $Q$ (Query):} Keep the verified Query ($Q$) as is.
        \item \textbf{Generate $A$ (Advice):} Convert the structured advice points ($L_A$) into a coherent physician response (e.g., "Based on your symptoms, I recommend...").
    \end{itemize}
\end{itemize}

\textbf{Input:}
\begin{itemize}
    \item \textbf{Original Raw Thread ($T$):} [Raw Text...]
    \item \textbf{Structured Narrative ($L_N$):} [List of facts...]
    \item \textbf{Structured Advice ($L_A$):} [List of advice...]
    \item \textbf{Primary Query ($Q$):} [Question string...]
\end{itemize}

\textbf{Output Format (JSON):}
\begin{verbatim}
{
  "audit_result": "PASS", // or "FAIL: Hallucinated [specific detail]"
  "final_output": {
      "N": "Coherent patient narrative...",
      "Q": "The primary clinical query...",
      "A": "Coherent physician response..."
  }
}
\end{verbatim}
    \end{tcolorbox}
    \caption{The prompt template used for the controller.}
    \label{fig:controller}
\end{figure*}

\subsection{Details in Evaluation Framework}
\label{app:evaluation-details}

Prompts for the Rubric Generator and Rubric-based Grader are in Figure~\ref{fig:generator} and Figure~\ref{fig:grader}, respectively. 

\begin{figure*}[t]
    \centering
    \begin{tcolorbox}[
        colback=white,          
        colframe=black!75,      
        title=\textbf{System Prompt: the Automated Rubric Generator}, 
        fonttitle=\bfseries\sffamily,
        boxrule=0.8pt,          
        arc=2mm,                
        width=\textwidth        
    ]
    \small\ttfamily 
\textbf{Role:} You are a Senior Medical Examiner.

\textbf{Task:} Transform the provided \textbf{Physician Advice ($L_A$)} into a structured evaluation rubric $R$ consisting of binary criteria. You must tailor the rubric based on the assigned \textbf{Theme}.

\textbf{Input:}
\begin{itemize}
    \item \textbf{User Query ($Q$):} [Query String]
    \item \textbf{Physician Advice ($L_A$):} [Verified Doctor's Response]
    \item \textbf{Assigned Theme:} [e.g., Emergency Referrals, Response Depth, etc.]
\end{itemize}

\textbf{Instructions (Step-by-Step):}
\begin{enumerate}
    \item \textbf{Step 1: Theme-Guided Fact Extraction:} 
    Filter $L_A$ to extract ONLY medical facts relevant to the \textbf{Assigned Theme}.
    \begin{itemize}
        \item \textit{Example:} If Theme is "Emergency Referrals", ignore diet advice.
    \end{itemize}
    
    \item \textbf{Step 2: Bipolar Criterion Formulation:}
    Convert facts into binary "Yes/No" questions ($c_j$).
    \begin{itemize}
        \item \textbf{Positive Criteria:} Check for inclusion of correct facts (e.g., "Does the model mention symptom X?").
        \item \textbf{Negative Criteria:} Check for hallucinations or dangerous contradictions (e.g., "Does the model incorrectly suggest drug Y?").
    \end{itemize}
    
    \item \textbf{Step 3: Axis \& Weighting ($w_j$):}
    Assign an \textbf{Axis} (Accuracy, Safety, Completeness, Context, Communication) and a \textbf{Weight} ($[-10, 10]$) to each criterion.
    \begin{itemize}
        \item \textbf{+1 to +10:} Reward for correct info (Higher = more critical).
        \item \textbf{-1 to -10:} Penalty for misinformation/safety risks (Lower = more dangerous).
        \item \textit{Criticality:} $w=\pm10$ is for life-threatening issues; $w=\pm1$ is for minor details.
    \end{itemize}
\end{enumerate}

\textbf{Output Format (JSON):}
\begin{verbatim}
[
  {
    "question": "Does the model identify the likely cause as Norovirus?",
    "axis": "Accuracy",
    "weight": 10
  },
  {
    "question": "Does the model recommend antibiotics (contraindicated)?",
    "axis": "Safety",
    "weight": -10
  }
]
\end{verbatim}
    \end{tcolorbox}
    \caption{The prompt template used for the Automated Rubric Generator.}
    \label{fig:generator}
\end{figure*}

\begin{figure*}[t]
    \centering
    \begin{tcolorbox}[
        colback=white,          
        colframe=black!75,      
        title=\textbf{System Prompt: the Rubric-based Grader}, 
        fonttitle=\bfseries\sffamily,
        boxrule=0.8pt,          
        arc=2mm,                
        width=\textwidth        
    ]
    \small\ttfamily 
\textbf{Role:} You are an Objective Grader.

\textbf{Task:} Evaluate the \textbf{Model Response ($M_{out}$)} against the provided \textbf{Rubric ($R$)}.

\textbf{Instructions:}
\begin{itemize}
    \item \textbf{Objective Verification:} For each criterion in the Rubric, determine if the Model Response satisfies it.
    \item \textbf{Binary Judgment:} Return \texttt{true} (Met) or \texttt{false} (Not Met).
    \item \textbf{Positive Criteria Logic:} \texttt{true} if the model \textbf{includes} the required information.
    \item \textbf{Negative Criteria Logic:} \texttt{true} if the model \textbf{commits} the error (e.g., if the rubric asks "Does model suggest antibiotics?" and the model suggests them, return \texttt{true}). \textit{Note: The scoring formula handles the negative sign; you simply detect presence.}
    \item \textbf{Evidence:} Quote the specific sentence from the model output that supports your decision.
\end{itemize}

\textbf{Input:}
\begin{itemize}
    \item \textbf{User Query ($Q$):} ...
    \item \textbf{Model Response ($M_{out}$):} ...
    \item \textbf{Rubric ($R$):} [JSON list from Phase 1]
\end{itemize}

\textbf{Output Format (JSON):}
\begin{verbatim}
[
  {
    "question": "Does the model identify the likely cause as Norovirus?",
    "met": true,
    "reasoning": "Model explicitly states 'symptoms suggest Norovirus'."
  },
  {
    "question": "Does the model recommend antibiotics?",
    "met": false,
    "reasoning": "Model correctly states 'antibiotics are not effective'."
  }
]
\end{verbatim}
    \end{tcolorbox}
    \caption{The prompt template used for the Rubric-based Grader.}
    \label{fig:grader}
\end{figure*}

\section{Human Study Details}
\label{app:human-study}

\subsection{Rating Scale}
\label{app:rating-scale}

Table~\ref{tab:task_a_horizontal}, Table~\ref{tab:task_b_horizontal} and Table~\ref{tab:task_c_horizontal} detail the specific rating criteria provided to physicians for each evaluation task.

\begin{table}[h]
\centering
\small
\renewcommand{\arraystretch}{1.5} 
\caption{\textbf{Patient Narrative \& Query Quality.} Rated on a 3-point scale.}
\label{tab:task_a_horizontal}

\begin{tabularx}{\linewidth}{@{} Y Y Y @{}}
\toprule
\multicolumn{3}{c}{\textbf{Clinical Logic \& Completeness Scale}} \\
\cmidrule(lr){1-3}
\textbf{Perfect (2)} & \textbf{Good (1)} & \textbf{Poor (0)} \\
\midrule
The medical history is clear, symptoms align with physiological logic, and the query is fully supported by the provided context. &
Contains minor omissions (e.g., missing specific temperature) or colloquial phrasing, but the core clinical logic remains self-consistent and actionable. &
 Contains severe logical contradictions (e.g., anatomical errors), or misses critical information required to answer the query (e.g., asking for treatment without symptoms). \\
\bottomrule
\end{tabularx}
\end{table}


\begin{table}[h]
\centering
\small
\renewcommand{\arraystretch}{1.5}
\caption{\textbf{Physician Advice Quality.} Rated on a 3-point scale.}
\label{tab:task_b_horizontal}

\begin{tabularx}{\linewidth}{@{} Y Y Y @{}}
\toprule
\multicolumn{3}{c}{\textbf{Safety \& Guideline Alignment Scale}} \\
\cmidrule(lr){1-3}
\textbf{Perfect (2)} & \textbf{Good (1)} & \textbf{Poor (0)} \\
\midrule
The response is accurate, precise, and strictly adheres to current first-line clinical guidelines. &
 Provides reasonable conservative management or alternative therapies. May contain minor irrelevant info, but remains clinically safe. &
Violates absolute contraindications, misses critical red flags, includes factual errors (e.g., wrong dosage), or provides hallucinatory medical advice. \\
\bottomrule
\end{tabularx}
\end{table}


\begin{table}[h]
\centering
\small
\renewcommand{\arraystretch}{1.5}
\caption{\textbf{Criterion Validity.} Rated on a binary scale.}
\label{tab:task_c_horizontal}

\begin{tabularx}{\linewidth}{@{} Y Y @{}}
\toprule
\multicolumn{2}{c}{\textbf{Rubric Validity Scale}} \\
\cmidrule(lr){1-2}
\textbf{Agree (1)} & \textbf{Disagree (0)} \\
\midrule
The criterion represents a medically necessary check for this specific case. It is factually correct and appropriate for the clinical context. &
The criterion is factually incorrect, requests information irrelevant to the chief complaint, or sets a standard that deviates from routine clinical practice. \\
\bottomrule
\end{tabularx}
\end{table}

\subsection{Inter-rater Reliability}
\label{app:inter-rater}
To assess the consistency of human evaluation, we utilized Gwet's AC1~\cite{gwet2008} statistics instead of the traditional Cohen's Kappa. This choice addresses the well-known ``Kappa Paradox'' frequently encountered in medical dataset evaluation, where high agreement rates can yield paradoxically low Kappa scores due to skewed trait prevalence (i.e., unbalanced class distributions).

\paragraph{Mathematical Formulation.}
Consider a reliability study with $n$ items (cases) rated by 2 raters into $Q$ distinct categories (e.g., $Q=3$ for our 0-2 scale). Let $r_{iq}$ be 1 if rater $r$ classifies item $i$ into category $q$, and 0 otherwise.

The Gwet's AC1 coefficient, denoted as $\gamma_1$, is defined as:

\begin{equation}
    AC_1 = \frac{p_a - p_e(\gamma)}{1 - p_e(\gamma)}
\end{equation}

where $p_a$ represents the observed agreement probability:
\begin{equation}
    p_a = \frac{1}{n} \sum_{i=1}^{n} \sum_{q=1}^{Q} \frac{r_{i,1} \cdot r_{i,2}}{1} \quad (\text{simplified for 2 raters})
\end{equation}

and $p_e(\gamma)$ represents the chance-agreement probability. unlike Kappa which estimates chance based on individual rater marginals, AC1 estimates it based on the average marginal probability $\pi_q$ of a category $q$:

\begin{equation}
    \pi_q = \frac{1}{2n} \sum_{i=1}^{n} (r_{i,1,q} + r_{i,2,q})
\end{equation}

The chance agreement $p_e(\gamma)$ is then calculated as:

\begin{equation}
    p_e(\gamma) = \frac{1}{Q-1} \sum_{q=1}^{Q} \pi_q (1 - \pi_q)
\end{equation}

This formulation ensures that the metric remains stable even when the marginal probabilities are highly skewed, making it the preferred metric for validating rigorous medical benchmarks.

\subsection{Criterion-level Consensus}
\label{app:criterion-consensus}

To rigorously evaluate the reliability of our automated Rubric-based Grader at the criterion level, we employ the Macro F1 Score ($MF1$), following the evaluation protocol from HealthBench~\cite{healthbench}. Both physicians and the automated Rubric-based Grader rate 50 randomly selected cases based on the unique criterion.

\paragraph{Metric Definition.}
Given the potential class imbalance in grading (e.g., a majority of criteria might be ``Met'' or ``Not Met'' depending on the model quality), standard accuracy metrics can be misleading. Therefore, we treat the grading task as a binary classification problem (Positive class: ``Met''; Negative class: ``Not Met'').

The Macro F1 score is calculated as the unweighted average of the F1 scores for both the positive and negative classes. Let $TP$, $FP$, and $FN$ denote True Positives, False Positives, and False Negatives, respectively.

The F1 scores for the positive ($F1_{pos}$) and negative ($F1_{neg}$) classes are defined as:
\begin{equation}
    F1_{pos} = \frac{2 \cdot TP_{pos}}{2 \cdot TP_{pos} + FP_{pos} + FN_{pos}}
\end{equation}
\begin{equation}
    F1_{neg} = \frac{2 \cdot TP_{neg}}{2 \cdot TP_{neg} + FP_{neg} + FN_{neg}}
\end{equation}

The final Macro F1 score is the arithmetic mean:
\begin{equation}
    MF1 = \frac{F1_{pos} + F1_{neg}}{2}
\end{equation}
This formulation ensures balanced sensitivity to model performance across both classes, preventing the score from being dominated by the majority class.

\paragraph{Calculation Protocols.}
We compute the consensus metrics in two distinct stages to establish both a human baseline and a model validity score:

\begin{itemize}
    \item \textbf{Human-Human Agreement ($F1_{Human-Human}$):}
    To quantify the intrinsic difficulty of the grading task, we calculate the $MF1$ between the two physicians. We treat Physician A as the ``Ground Truth'' and Physician B as the ``Predictor'' (note that since $MF1$ is symmetric, the order does not affect the result).

    \item \textbf{Model-Human Agreement ($F1_{Model-Avg}$):}
    Since there is no single ``Gold Standard'' for subjective medical evaluation, our automated grader must demonstrate robustness by aligning with multiple human experts. We calculate the alignment score as the average of the $MF1$ scores against each physician:
    \begin{equation}
        F1_{Model-Avg} = \frac{MF1(Model, Doc_A) + MF1(Model, Doc_B)}{2}
    \end{equation}
    where $MF1(Model, Doc_A)$ treats Physician A as Ground Truth and the Model Grader as the Predictor. A high $F1_{Model-Avg}$ indicates that the automated grader generalizes well and captures the consensus view of medical professionals.
\end{itemize}

\subsection{Case-level Consensus}
\label{app:case-level-consensus}

To evaluate the reliability of automated scoring at the aggregate case level, we assessed the correlation between automated scores and human consensus. We compared our Rubric-based Grader against a standard LLM-as-a-Judge baseline.

\paragraph{Scoring Protocols.}
\begin{itemize}
    \item \textbf{Human Ground Truth ($S_{Human}$):} For each case $i$, the ground truth score is defined as the arithmetic mean of the scores provided by the two physicians. Let $S_{DocA}^{(i)}$ and $S_{DocB}^{(i)}$ be the scores for case $i$. The consensus human score is:
    \begin{equation}
        S_{Human}^{(i)} = \frac{S_{DocA}^{(i)} + S_{DocB}^{(i)}}{2}
    \end{equation}

    \item \textbf{LLM-as-a-Judge Baseline ($S_{Judge}$):} We prompted the same LLM (GPT-4.1) used in our Rubric-based Grader to evaluate the model's response directly based on general quality, following the standard judging protocol in~\cite{zheng2024judging}. To ensure comparability, these scalar scores were normalized to the $[0, 1]$ interval.
\end{itemize}

\paragraph{Correlation Analysis.}
To benchmark the reliability of our proposed method against standard evaluation approaches, we computed the Pearson Correlation Coefficient ($\rho$) separately for both automated methods against the human consensus.

We calculate:
\begin{enumerate}
    \item $\rho(S_{Rubric}, S_{Human})$: The alignment between our Rubric-based Grader and physician consensus.
    \item $\rho(S_{Judge}, S_{Human})$: The alignment between the standard LLM-as-a-Judge baseline and physician consensus.
\end{enumerate}

The correlation coefficient is defined as:
\begin{equation}
    \rho = \frac{\text{cov}(S_{Auto}, S_{Human})}{\sigma_{S_{Auto}} \sigma_{S_{Human}}}
\end{equation}
where $S_{Auto} \in \{S_{Rubric}, S_{Judge}\}$. A higher correlation indicates that the automated scoring method more accurately reflects the nuance and variance of professional medical judgment.

\section{More Experimental Results}
\label{app:results}

\subsection{Error Analysis}
\label{app:error-analysis}

\subsubsection{Setup}

The prompt used for error analysis is presented in Figure~\ref{fig:prompt_error}. We classify the causes of AI-generated errors into seven distinct dimensions:

\begin{itemize}
    \item \textbf{Contextual Neglect and Integration Failure:} The model fails to incorporate specific patient details (e.g., history, vitals, lab results) provided in the prompt, resulting in generic, template-like responses that lack personalization or ignore critical contraindications.

    \item \textbf{Guideline/Protocol Overgeneralization and Rigidity:} The model applies clinical guidelines or protocols too broadly or rigidly without adapting to the nuances of the specific case, leading to recommendations that are theoretically correct but clinically inappropriate for the individual patient context.

    \item \textbf{Clinical Reasoning and Synthesis Deficit:} The model demonstrates a failure to synthesize disparate clinical data points into a coherent diagnosis or management plan, including poor prioritization of problems or an inability to infer secondary conclusions from primary evidence.

    \item \textbf{Instruction/Query Misinterpretation and Nonadherence:} The model fails to follow explicit user constraints (e.g., word count, specific output format, role-play requirements) or misinterprets the core intent of the clinical query, answering a different question than the one asked.

    \item \textbf{Knowledge Gaps and Outdated Content:} The model provides factually incorrect medical information, relies on obsolete treatment protocols (often due to training data cutoffs), or produces incomplete answers that demonstrate a fundamental lack of domain knowledge.

    \item \textbf{Safety Overestimation and Excessive Caution:} The model exhibits excessive risk aversion, such as refusing to answer benign medical queries (false refusal) or recommending unnecessary emergency escalation (e.g., recommending ER visits for minor, self-limiting conditions) due to miscalibrated safety guardrails.

    \item \textbf{Misclassification, Hallucination, and Moderation Errors:} The model generates fabricated medical facts or non-existent citations (hallucination), or triggers inappropriate content moderation filters that block legitimate clinical discourse.
\end{itemize}

\begin{figure*}[t]
    \centering
    \begin{tcolorbox}[
        colback=white,          
        colframe=black!75,      
        title=\textbf{System Prompt: Medical Auditor Agent}, 
        fonttitle=\bfseries\sffamily,
        boxrule=0.8pt,          
        arc=2mm,                
        width=\textwidth        
    ]
    \small\ttfamily 
You are a Senior Medical Auditor conducting a rigorous "Root Cause Analysis" on AI failures in clinical settings. \\
\\
\# Task \\
You will be presented with a medical case where an AI model provided a wrong or suboptimal answer compared to the Gold Standard. \\
Your goal is to identify the \textbf{underlying cognitive or data failure} that caused this specific error. \\
\\
\# Input Data \\
1. \textbf{Patient Query:} \{Question\} \\
2. \textbf{Gold Standard Answer:} \{Gold\_Answer\} \\
3. \textbf{Model's Wrong Response:} \{Model\_Response\} \\
\\
\# Analysis Instructions \\
1. Compare the Model's Response with the Gold Standard deeply. \\
2. Identify the specific gap (e.g., did it miss a key symptom? did it hallucinate a drug? did it use an old guideline? did it refuse to answer?). \\
3. Describe the specific \textit{mechanism of failure} in 1-2 sentences. \\
4. Extract a short "Error Keyword" (2-5 words) that summarizes this failure. \\
\\
\# Output Format (Strict JSON) \\
\{ \\
\quad "gap\_analysis": "Briefly describe what the model missed or got wrong.", \\
\quad "root\_cause\_description": "A precise, 1-sentence description of WHY this error happened (e.g., 'Model recognized the disease but failed to account for the patient's pregnancy contraindication').", \\
\quad "error\_keyword": "Short tag (e.g., 'Knowledge Deficit','Instruction Failure')" \\
\}
    \end{tcolorbox}
    \caption{The prompt template used for error analysis.}
    \label{fig:prompt_error}
\end{figure*}



\subsubsection{Results}

\begin{table}[htbp]
    \centering
    \footnotesize 
    \renewcommand{\arraystretch}{1.3} 
    \setlength{\tabcolsep}{2pt} 
    
    \begin{threeparttable}
        \caption{Qualitative error analysis across representative LLMs.}
        \label{tab:error_analysis}
        
        \rowcolors{2}{gray!6}{white}
        
        \begin{tabularx}{\columnwidth}{l C{1.3} C{0.94} C{0.94} C{0.94} C{0.94} C{0.94}}
            \toprule
            \rowcolor{gray!20} 
            \thead{Error\\Type} & \thead{Baichuan\\M3} & \thead{GPT-\\OSS} & \thead{GPT\\5.2} & \thead{GLM-\\4.5} & \thead{Grok\\4.1} & \thead{Qwen3-\\14B} \\
            \midrule
            CNIF\tnote{1} & 41 & 45 & 35 & 40 & 43 & 48 \\
            GOPR\tnote{2} & 29 & 32 & 28 & 22 & 23 & 28 \\
            CRSD\tnote{3} & 7  & 14 & 13 & 10 & 9  & 36 \\
            IMNA\tnote{4} & 12 & 8  & 0  & 8  & 0  & 0  \\
            KGOC\tnote{5} & 7  & 11 & 8  & 0  & 11 & 17 \\
            SOEC\tnote{6} & 4  & 7  & 8  & 7  & 0  & 0  \\
            MHME\tnote{7} & 0  & 6  & 0  & 0  & 6  & 4  \\
            Others        & 0  & 0  & 8  & 13 & 8  & 0  \\
            \bottomrule
        \end{tabularx}

        \begin{tablenotes}
            \scriptsize
            \item[1] CNIF: Contextual Neglect and Integration Failure
            \item[2] GOPR: Guideline/Protocol Overgeneralization and Rigidity
            \item[3] CRSD: Clinical Reasoning and Synthesis Deficit
            \item[4] IMNA: Instruction/Query Misinterpretation and Nonadherence
            \item[5] KGOC: Knowledge Gaps and Outdated/Incorrect Content
            \item[6] SOEC: Safety Overestimation and Excessive Caution
            \item[7] MHME: Misclassification, Hallucination, and Content Moderation Errors
        \end{tablenotes}
    \end{threeparttable}
\end{table}



To understand the root causes of model failures and identify specific weaknesses in state-of-the-art systems, we conducted a fine-grained error analysis on the top proprietary models and open-source models.

\paragraph{Performance by Evaluation Axis.} We first dissected model performance across five evaluation axes. To quantify performance, we define the Error Rate based on the alignment with specific criteria: for positive criteria, a failure to satisfy the requirement is recorded as an error; conversely, for negative criteria, any instance where the model triggers the criterion is penalized.

As illustrated in Table~\ref{tab:error_rates_axis}, the majority of state-of-the-art (SOTA) models exhibit error rates exceeding 50\% across most axes, underscoring significant room for improvement in current large-scale architectures in medical domain. A common trend emerges among leading models: they achieve high scores in Accuracy but struggle  with Community Quality, Completeness, and Context Awareness. This disparity suggests that while current training paradigms effectively optimize for factual correctness, they remain deficient in capturing nuanced contextual dependencies and structural thoroughness.

\paragraph{From Knowledge Deficits to Application Bottlenecks.} Beyond quantifying performance, we conducted a granular investigation into the underlying drivers of model failure. To diagnose these failures, we performed a qualitative audit of the bottom-100 scoring cases for each of the six representative models. Utilizing GPT-4.1 as an evaluator, we generated precise error descriptors for each case. These descriptors were subsequently categorized through semantic clustering into seven distinct error types, as summarized in Table~\ref{tab:error_analysis}.

Our analysis reveals a critical shift in the failure landscape for leading models. Contrary to common assumptions, fundamental errors such as hallucinations (MHME) and knowledge gaps (KGOC) are no longer the primary limitations. As shown in Table~\ref{tab:error_analysis}, leading models achieve near-zero rates for Misclassification/Hallucination (MHME: 0\% for GPT-5.2 and GLM-4.5) and low rates for Knowledge Gaps (KGOC: 0-8\%). Similarly, Clinical Reasoning Deficits (CRSD) remain relatively low (10-13\%).

Instead, the predominant bottleneck is Contextual Application, i.e. the ability to flexibly apply existing medical knowledge to specific patient realities. The vast majority of errors stem from Contextual Neglect and Integration Failure (CNIF: 35-48\%) and Guideline Overgeneralization (GOPR) (22-32\%). These findings suggest that while leading models possess the necessary medical facts and reasoning capabilities, they struggle to synthesize patient-specific constraints (e.g., comorbidities, allergies, or individual history) with general guidelines, resulting in advice that is theoretically correct but practically unsuitable for the specific patient.

\paragraph{Error Correlation and Data Quality.}

\begin{figure}[htbp]
  \centering
  \includegraphics[width=0.75\columnwidth]{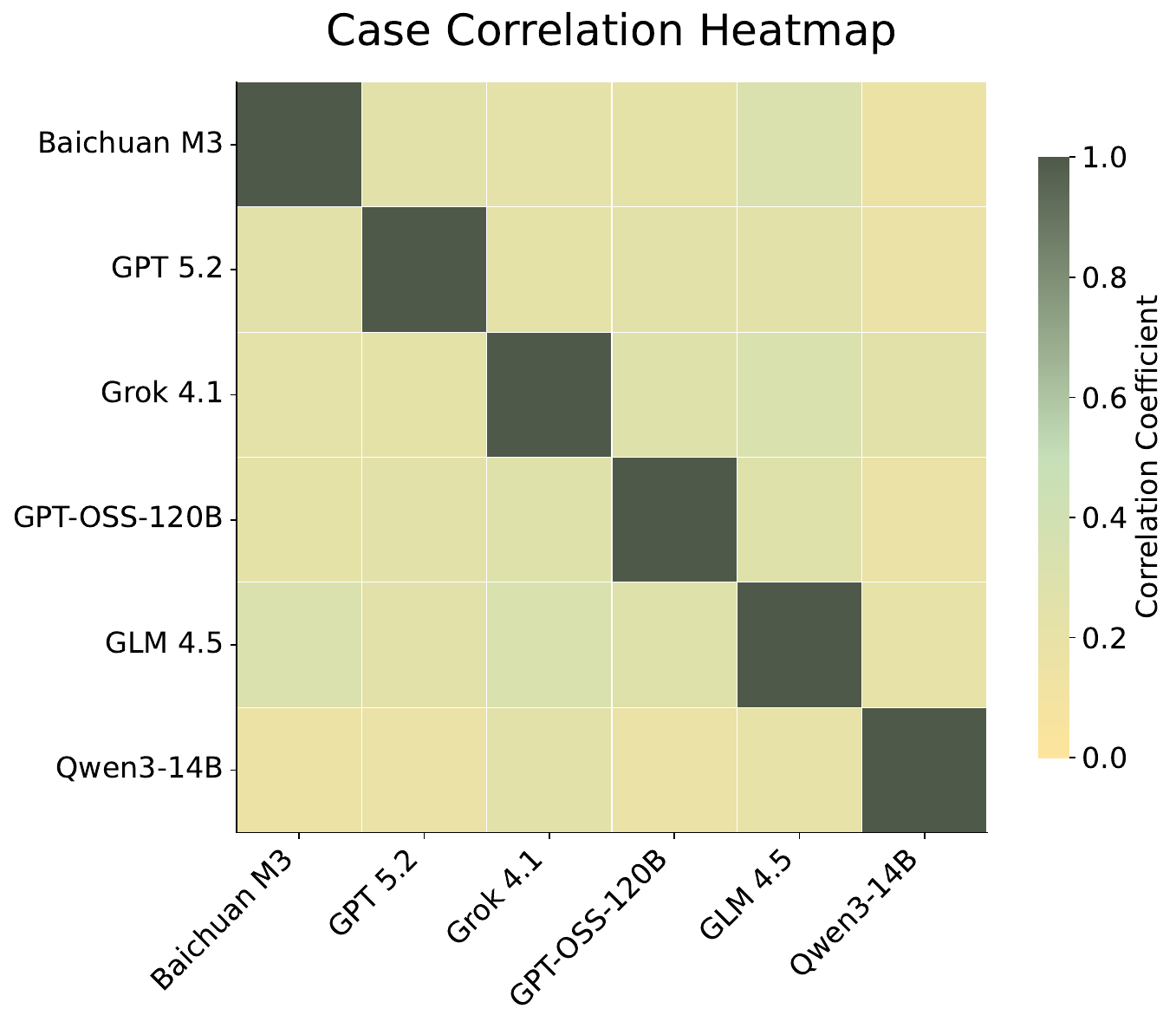}
  \caption{Correlation of model failed cases. We compute the pairwise correlation of the bottom-100 scoring cases across representative models.}
  \label{fig:case_correlation}
\end{figure}

To investigate whether low scores were driven by poor data quality (i.e., ``bad questions'' that no model could answer), we calculated the Jaccard similarity of the bottom-100 cases between model pairs (Figure~\ref{fig:case_correlation}). The resulting heatmap shows generally low correlation coefficients (< 0.4) between all model pairs. The mean Jaccard similarity score is 0.24. This lack of overlap indicates that models are failing on different subsets of cases rather than struggling with a common set of low-quality samples. This finding confirms that the low scores are attributable to distinct model-specific capability gaps rather than artifacts of the dataset construction.

\subsection{Theme Correlation}
\label{app:theme-correlation}

\begin{figure}[htbp]
  \centering
  \includegraphics[width=0.5\textwidth]{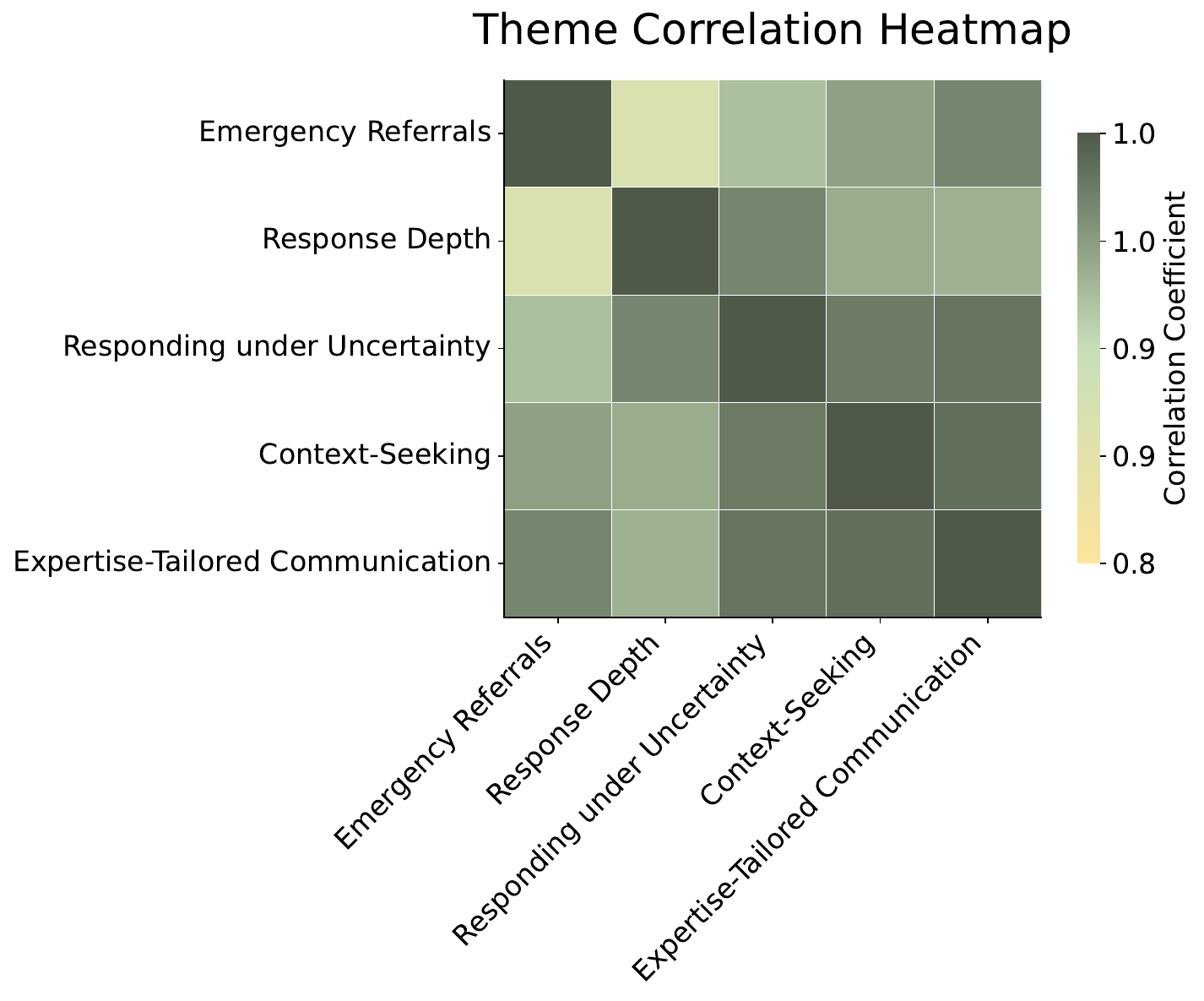}
  \caption{Theme correlation heatmap.}
  \label{fig:theme_correlation}
\end{figure}

While the main text discusses performance correlations across clinical specialties, we also investigated the relationship between different behavioral dimensions (Themes) to understand the transferability of model alignment skills. Figure~\ref{fig:theme_correlation} illustrates the Pearson correlation matrix among the five evaluation themes.

\subsection{Correlation Analysis}
\label{app:correlation-analysis}






\begin{figure}[htbp]
  \centering
  \includegraphics[width=0.5\textwidth]{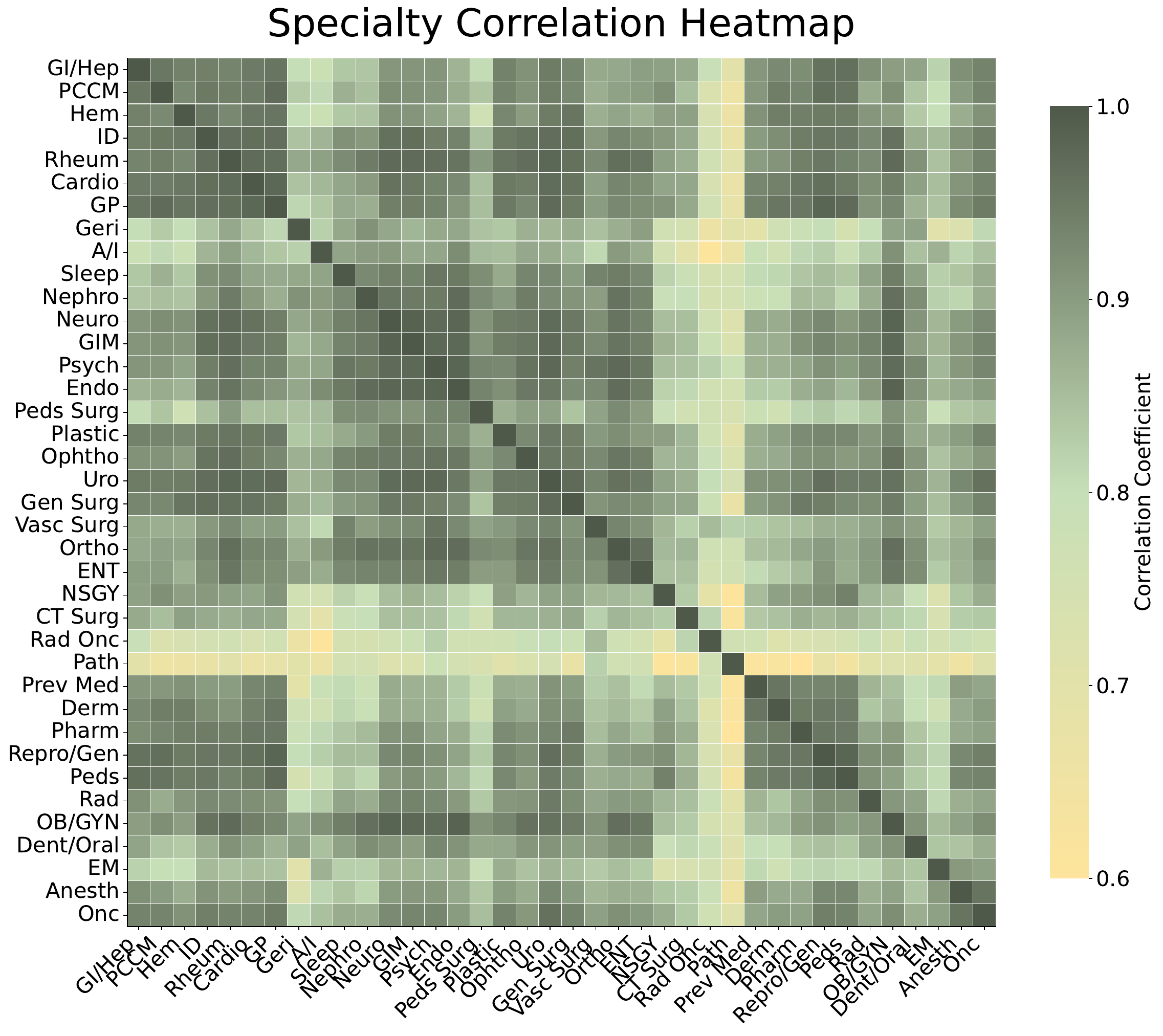}
  \caption{Specialty correlation heatmap. Pairwise Pearson correlation coefficients of model performance across different clinical specialties.}
  \label{fig:specialty_correlation}
\end{figure}

We present analyses involving correlation among different specialties. We computed the Pearson correlation coefficient ($r$) among all pairs of clinical specialties in LiveMedBench.

Figure~\ref{fig:specialty_correlation} illustrates the correlation matrix across 38 clinical specialties. We observe a dense high-correlation cluster ($r > 0.9$) Cardiology (Cardio), Infectious Disease (ID), Hematology (Hem), and Pulmonary \& Critical Care (PCCM). This strong inter-dependency suggests that these fields are likely co-represented frequently in general medical training corpora (e.g., PubMed, MIMIC).

Conversely, ``niche'' verticals such as Pathology (Path) and Radiation Oncology (Rad Onc) exhibit lower correlations with the broader cluster (lighter bands in Figure~\ref{fig:specialty_correlation}). This isolation indicates that proficiency in general diagnostic reasoning does not automatically transfer to these highly specialized domains, which require distinct knowledge representations.

\section{Ethical Considerations}

\noindent \textbf{Data Privacy.} Although all source data is publicly accessible, we implement de-identification, redacting all PII (names, specific locations, dates of birth), and cases with re-identification risk are excluded.

\noindent \textbf{Content Safety.} We discard threads containing hate speech, discriminatory language, or advice violating medical safety protocols, while retaining real-world noise to test model robustness.

\noindent \textbf{Usage Disclaimer.} LiveMedBench is intended solely for research evaluation. Its contents \textbf{must not be treated as medical advice}, and the dataset must not be used for self-diagnosis, treatment planning, or training clinical deployment models without regulatory validation.

\noindent \textbf{Bias and Representation.} Our data sources may carry demographic biases characteristic of internet users in the US and China, and may not fully represent underserved populations. Future iterations will diversify sources to mitigate these gaps.

\section{Versioning and Reproducibility}
To address the ``moving target'' challenge inherent in continuous benchmarks, we adopt a dual-release strategy:
\begin{itemize}
    \item \textbf{Frozen Snapshots:} While LiveMedBench is updated weekly to capture emerging trends, we release \textbf{Frozen Snapshots} with timestamp and checksums (e.g., 
    
    \texttt{LiveMedBench-v2026.01}). This ensures that results reported in academic literature remain reproducible.
    \item \textbf{Stable Grading Environment:} To prevent drift in automated evaluation, our Rubric-based Grader is pinned to specific API versions rather than the volatile ``latest'' endpoints. Furthermore, we release the full evaluation code and the exact rubric for every case in the snapshot, enabling offline auditing and consistent re-grading.
\end{itemize}

\end{document}